\documentclass[default,iicol,pdflatex]{sn-jnl} 
\jyear{2021}%
\theoremstyle{thmstyleone}%
\usepackage{dblfnote}
\theoremstyle{thmstyletwo}%
\usepackage[section]{placeins}
\theoremstyle{thmstylethree}%
\begin{document}
\title[]{Classification of news spreading barriers}

\author*[1,2]{\fnm{Abdul} \sur{Sittar}}\email{abdul.sittar@ijs.si}

\author[1,2]{\fnm{Dunja} \sur{Mladeni\'{c}}}\email{dunja.mladenic@ijs.si}

\author[2]{\fnm{Marko} \sur{Grobelnik}}\email{marko.grobelnik@ijs.si}

\affil*[1]{\orgname{Jozef Stefan International Postgraduate School}, \orgaddress{\street{Jamova Cesta 39}, \city{Ljubljana}, \postcode{1000}, \country{Slovenia}}}

\affil[2]{\orgdiv{Department for Artificial Intelligence}, \orgname{Jozef Stefan Institute}, \orgaddress{\street{Jamova Cesta 39}, \city{Ljubljana}, \postcode{1000}, \country{Slovenia}}}


\abstract{
News media is one of the most effective mechanisms for spreading information internationally, and many events from different areas are internationally relevant. However, news coverage for some news events is limited to a specific geographical region because of information spreading barriers, which can be political, geographical, economic, cultural, or linguistic. In this paper, we propose an approach to barrier classification where we infer the semantics of news articles through Wikipedia concepts. To that end, we collected news articles and annotated them for different kinds of barriers using the metadata of news publishers. Then, we utilize the Wikipedia concepts along with the body text of news articles as features to infer the news-spreading barriers. We compare our approach to the classical text classification methods, deep learning, and transformer-based methods. The results show that the proposed approach using Wikipedia concepts based semantic knowledge offers better performance than the usual for classifying the news-spreading barriers.}


\maketitle

\section{Introduction}\label{sec:Introduction}
Media coverage of local and global events defines and limits the discourse associated with different events. The priority is given to different contents based on cultural, political, social, linguistic, geographical, and economic biases \cite{ref:miles2007role, ref:archetti2008news}. Similarly, the news relating to local events involves domestic factors, whereas the news about global events involves national and international factors that affect their news flow. These factors again include economic, political, cultural, linguistic, and geographical influences as \cite{ref:sittar2022analysis} concluded that depending on the nature of an event, there are variations in information-spreading behavior across the different barriers including economic, cultural, geographical, political, and linguistic. Classification of these barriers can be helpful in the context of numerous real-world applications, such as event-centric news analysis, suspicious news detection, and content recommendations to readers and subscribers. Thus, it is highly important to classify the barriers to massive news spreading related to different events.
\newline
\newline
It is important to understand the influence of the above-mentioned barriers to news spreading. Economic stability is one of the factors that influence media coverage \cite{ref:grasland2020international}. Moreover, the influence of economic power varies across different events and issues (e.g. protests, online privacy, disasters) \cite{ref:segev2015visible, ref:shahin2016right}.

\begin{figure*}
\centering
        \includegraphics[width=0.30\textwidth]{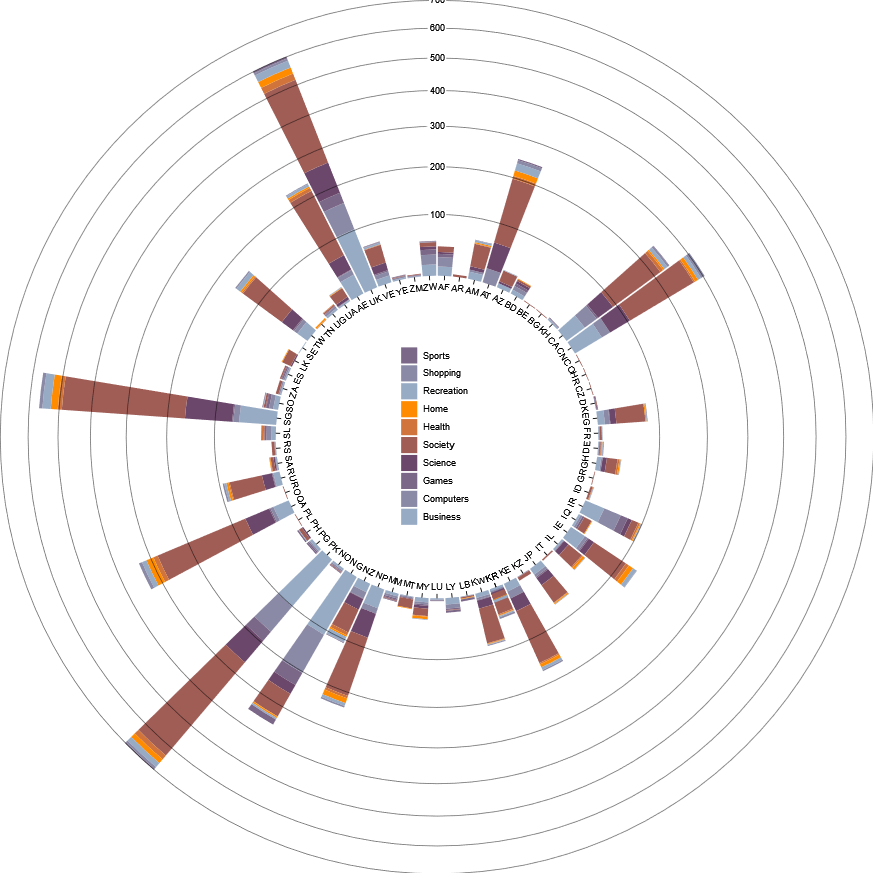}
        \includegraphics[width=0.30\textwidth]{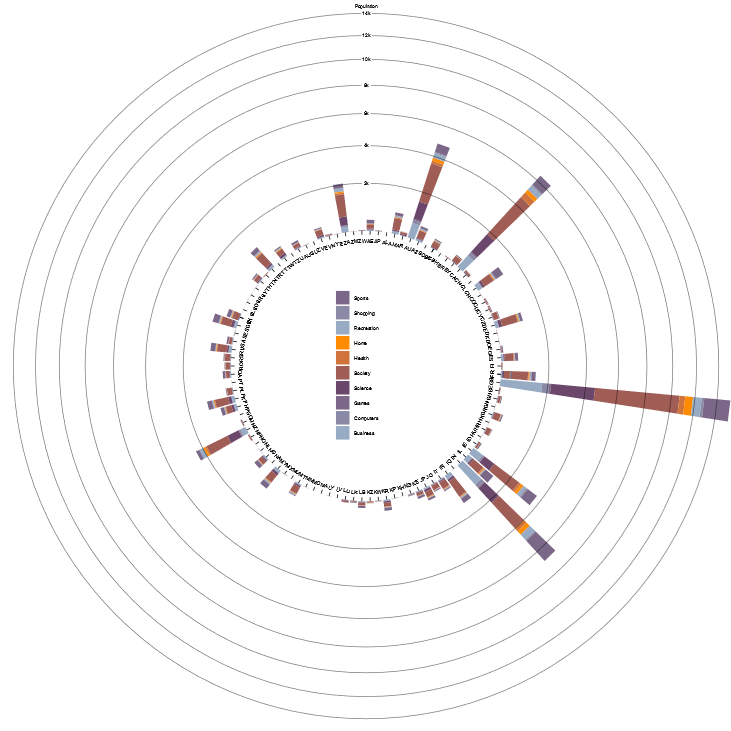}
        \includegraphics[width=0.30\textwidth]{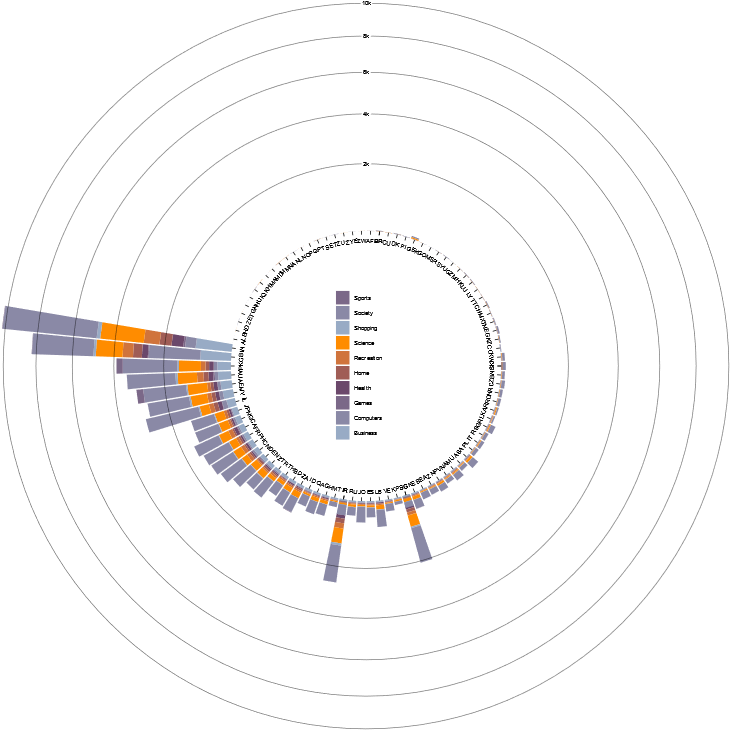}
        \caption{\bf The circular bar charts show the statistics about the news articles that have the labels "Information-crossing", "information-not-crossing", and "unsure" respectively (from left to right) for all the ten different categories. The circles show the count of the news articles, each bar represents a country, whereas the colors in each bar represent ten different categories (business, computers, games, and health, etc.). The purpose of this figure is to show the variations of the number of news articles that are either crossing or unsure or not crossing a barrier for different countries (see Section \ref{sec:datadescription})}. 
        \label{fig:totalfields}
\end{figure*}

 The magnitude of economic interactivity between countries can also impact the news flow \cite{ref:wu2007brave}. The national context in which the journalists work is frequently followed by news organizations. The SARS pandemic study, which discovered that cross-national contextual factors including political and economic situations affect news selection, is one of the related cases \cite{ref:camaj2010media}. Political ideology is another factor that influences media coverage and news spreading. Also one of the factors involved in producing fake news or rumors is the political effect \cite{ref:chen2020incentive, ref:koloski2022knowledge}. \cite{ref:hosni2020minimizing} presented a model to capture the spreading process of rumors on social networks.  A great amount of work regarding fake news dwells on different strategies and due to the engagement of journalists and political players, it has been convincingly demonstrated that controlling the news and making appropriate changes is a major method employed by news agencies \cite{ref:bakshy2015exposure, ref:maurer2018networking}. One of the determinants for influencing news spreading and coverage is the country's geographic and population size \cite{ref:wu2003homogeneity, ref:golan2009determinants}. According to certain theories, countries with close distances have some degree of cultural and linguistic affinities and because of that the flow of news spreading is much higher than in countries with long distances \cite{ref:wu2003homogeneity, ref:wu2007brave, ref:segev2014news, ref:segev2015visible, ref:erdmann2016machine}.
\newline
\newline
Generally, different types of semantic features have been used to perform news classification depending on the task \cite{ref:levi2019identifying, ref:rahmawati2016word2vec}. For instance, vectorized semantic and syntactical features for the spread of fake news over social, political, and economic context \cite{ref:kumar2022fake}, and semantic features like sentiment, entities or facts for fake news classification \cite{ref:bracsoveanu2019semantic}. Similarly, Stylistic and bag-of-word have been tested for the news classification at the publisher or regional level \cite{ref:sittar2022stylistic}. In this paper, we explore the classification of barriers to massive news spreading related to different events.  We are interested in exploring the variations in news spreading across different topics and different barriers. We focus on five different types of barriers including cultural, political, linguistic, economic, and geographic. Since the considered barriers deal at the international level, we assume that the Wikipedia concepts of news articles including entities (locations, people, organizations) or non-entities (things such as personal computers, and toys) will help in the classification of barriers.

\subsection{Motivation}
The motivations behind our work are stemmed from the following facts:
\begin{itemize}
\item  The news agencies/news publishers always want to have more viewership of their content to earn more money. A news article has mainly consisted of two things. Selection of words/terms to report about any event and selection of events to be reported in a news article. Then the result is subsequent news reporting on the same event by other publishers. During this news reporting, many barriers may stop it from spreading further. These barriers could be of these: political, geographical, economic, cultural, and linguistic. In this context, the barrier classification in news spreading is getting attention as an important research problem.

\item The barrier classification intends to assist newspapers in general, but can also be useful for the public. Researchers who want to know the reasons for cultural differences in different communities may learn by comparing the written news articles. Thus, developing an efficient and automatic barrier classification system for newspapers comes out as an essential task. To the best of our knowledge, there is a lack of studies that address this challenging task. 

\item By modeling the barriers (cultural, political, economic, geographic, and linguistic), news publishers can develop a better strategy to select an event and report about it, make models that take the news articles as input, and as a consequence control or modify reporting content, and in general, train systems to be better at detecting above mentioned barriers.

\end{itemize}

\subsection{Contributions}
The original scientific contributions of this paper are:

\begin{itemize}
\item A novel approach to barrier classification based on news meta-data.
\item An annotation process, and class definitions.
\item A novel approach to inferring the news spreading barriers using Wikipedia concept based semantic knowledge.
\end{itemize}

\subsection{Hypothesis and research questions}\label{subsec:hypothesis}
Barrier classification faces the challenge of efficiently analyzing huge amounts of news text. Our research hypothesis states that Wikipedia concept based semantic annotation of news articles will help in classifying the news-spreading barriers. We explore ten different types of news in this context including home, health, business, sports, recreation, shopping, computers, science, society, and games. In order to aid understanding of the influence of different barriers on different types of news, this article set three research questions:

\noindent\textbf{Q1:} Does the information spreading in news varies across different topics and different barriers?  \newline
\noindent\textbf{Q2:} What prominent relations appear between Wikipedia concepts and different barriers and categories?  \newline
\noindent\textbf{Q3:} Which classification methods (classical or deep learning methods) yield the best performance to barrier classification task?  \newline

The remainder of the paper is structured as follows. Section~\ref{sec:RelatedWork} describes the related work on an overview of the news spreading problems, the economic aspects of the news spreading, and breaching the barriers to extending viewership. The approach used for barrier classification is explained in Section~\ref{sec:approach}. The data collection and the annotation guidelines are presented in Section~\ref{sec:datadescription}. We present the experimental results in Section \ref{sec:ExpResults}. 
Section~\ref{sec:conclusions} concludes the paper and outlines the areas for future work.

\section{Related Word}\label{sec:RelatedWork}
In this literature review, we present different economic aspects connected with online news spreading, the cultural influence in the news spreading, and the role of content and the framing of news events by the news media. \newline

\noindent \textbf{Economic aspects connected with the online news spreading} 
Effective dissemination is the key to bridging the gap in information spreading. For the scientists and the practitioners, it is necessary to participate in explicit, accurate, and unbiased dissemination of their respective areas of expertise to the public \cite{ref:kelly2019spreading}. In the early stages of online experiments on the news spreading, there was fear that online content may erode the print edition. Therefore, the idea of charging a subscription fee to the users for online news access, and after that, the advertising model followed \cite{ref:chyi2002explorative}. Newspapers have always been very valuable advertising channels for promotional campaigns, e.g. couponing, retailer ads, etc., informative campaigns which provide extensive product information, and pure branding campaigns. Newspapers are a flexible medium that can reach large audiences although they can be used to address local targets. Newspapers are regarded as financially stable when 40-70 percent of their income comes from the advertising revenues \cite{ref:berte2008newspapers}. There are many issues and confusions about the profit of online news media. Although the number of online newspapers is increasing \cite{ref:tan2011digital}, whether this will become a financially successful business or not is still not clear \cite{ref:rahma2020impact}. Uncertainty exists over how online newspapers define important things: a market that spans the local and global levels, placement in the market, connection between online and print products, and establishment of key strategies. Because a market consists of both consumers and suppliers and because online practitioners are constantly experimenting with the new mediums, market research frequently focuses on user demographics. However, online publishers' perspectives are equally, if not more, important in understanding online newspaper economies \cite{ref:tan2011digital}. Online newspapers have experimented with various revenue models such as subscriptions, advertising, pay-per-use, sponsorships, web site development, serving as ISPs (Internet service providers), and e-commerce \cite{ref:chyi2000online, ref:franz2003customer, ref:awate2014survey}. These models define the geographical market for their online products. These models ask the following questions from participants - Do they define themselves as local, metro, regional, national, or global publications? Their response indicated a geographic market definition \cite{ref:ballon2014old}. Apart from the economic aspects of news spreading, media activities are a means to secure social, cultural, or political status \cite{ref:ballon2014old}.
\newline

\noindent\textbf{Cultural influence in the news spreading} 
The result of communication is not only situation-specific but also inherently culturally bound because it is entrenched in human acts with intentions, interests, and wants as well as larger institutional, social, and cultural systems \cite{ref:jiang2020relying}. A culture-specific ideology is defined as the values, beliefs, attitudes, or interests expressed in a source text that is associated with a particular culture or source and that may be viewed as undesirable or incompatible with the dominant values, beliefs, attitudes, or interests of another culture or subculture. It defines the strategies adopted by text producers in bridging the divides in global news transmission. According to MCNelly's theory, the more distance an intermediary communicator has to travel before learning about a news occurrence, the less personally invested he is in it and the more he considers its "marketability" to editors or readers \cite{ref:vuorinen1994crossing}. It has been said that countries with close distances share culture and the news reporting on the same events will not differ due to ideology, culture, and geopolitics \cite{ref:segev2015visible, ref:ma2017does}. Countries that share a common culture are expected to have heavier news flow between them when reporting on similar events \cite{ref:wu2007brave}. There are many quantitative studies that found demographic, psychological, socio-cultural, source, system, and content-related aspects \cite{ref:al2017impact}.
\newline

\noindent \textbf{Framing of news events by news media and role of content} 
The role of content is an essential research topic in news spreading. Media economics scholars especially showed their interest in a variety of content forms since content analysis plays a vital role in individual consumer decisions and political and economic interactions \cite{ref:fico2008content}. In content, a frame is a means to highlight certain elements of a seen reality in a communication text so as to support a specific problem definition, causal interpretation, moral assessment, and/or therapy proposal for the thing being described. There are four places where frames can be found during communication: the text, the recipient, the communicator, and the culture \cite{ref:reese2007framing}. The inverted pyramid reporting method, where the most significant facts are presented in order of importance, is a key component of news framing. Bias in the news can manifest in a variety of ways, these include "source bias", "unbalanced presentation of contested themes", and "frequent usage of packaged formula" \cite{ref:walter2019news}. Scheufele identifies five factors that influence how journalists frame news. These include societal expectations and ideals, organizational demands and restrictions, pressure from interest groups, journalistic practices, and journalists' ideological or political leanings \cite{ref:obijiofor2010press}. A vast body of literature exists on how the news media frame the news events and consequently influence public perception of those events \cite{ref:lamidi2016newspaper}. Existing literature posit that framing is often used intentionally for the purpose of changing the perception of content and to cater this, different computational methods have been applied \cite{ref:king2017news, ref:sheshadri2021detecting}.
\newline

\noindent\textbf{News classification methods}
Different text classification methods have been used to perform the classification of news articles belonging to different tasks \cite{ref:elnagar2020arabic, ref:samadi2021deep, ref:buvzic2018lyrics}. \cite{ref:kula2019application} presents a hybrid architecture connecting BERT with RNN and uses it to create models for detecting fake news. A fake news detection model using the n-gram analysis and classical machine learning techniques is proposed where SVM appears as the best classifier \cite{ref:ahmed2017detection}. It makes a comparison between two different feature extraction techniques and six different classical machine learning techniques. PAN is a series of scientific events and shared tasks which include classification based on textual data collected from social media \cite{ref:rangel2016overview, ref:alvarez2020author, ref:bevendorff2021overview}. 
\cite{ref:saleh2021opcnn} proposed novel approaches based on machine learning and deep learning for the fake news detection systems to address this phenomenon. It compares the performance of an optimized convolution neural network model with RNN, LSTM, and six regular ML techniques: Decision Tree, Logistic Regression, K Nearest Neighbor, Random Forest, SVM, and Naive Bayes using the four fake news benchmark datasets. 
\cite{ref:arora2022performance} applied these methods and feature engineering techniques such as count vectorizer, TF-IDF, and word2vec. It shows that multinomial Naive Bayes with count vectorizer outperforms Hindi news headlines related to different categories (entertainment, sports, tech, lifestyle). 
\newline
\newline
\noindent\textbf{Semantic knowledge for text classification}
Semantic knowledge is used to improve the performance of text mining algorithms by adding more semantic text \cite{ref:kiefer2022case, ref:bloehdorn2004boosting, ref:wang2013improving}. 
Different tasks utilize different types of semantic text such as knowledge graphs, WordNet, Open Directory Project, or Wikipedia \cite{ref:shanavas2021knowledge,ref:mansuy2006evaluating,ref:shin2017utilizing}. 
Wikipedia has been used for many studies as an external knowledge resource \cite{ref:mourino2018wikipedia, ref:poyraz2012exploiting, ref:hu2009exploiting}, we utilize the wikipedia concepts as a knowledge source for barrier classification.

\section{Approach}\label{sec:approach}
\begin{figure*}
\centering
\includegraphics[keepaspectratio=true,scale=0.9]{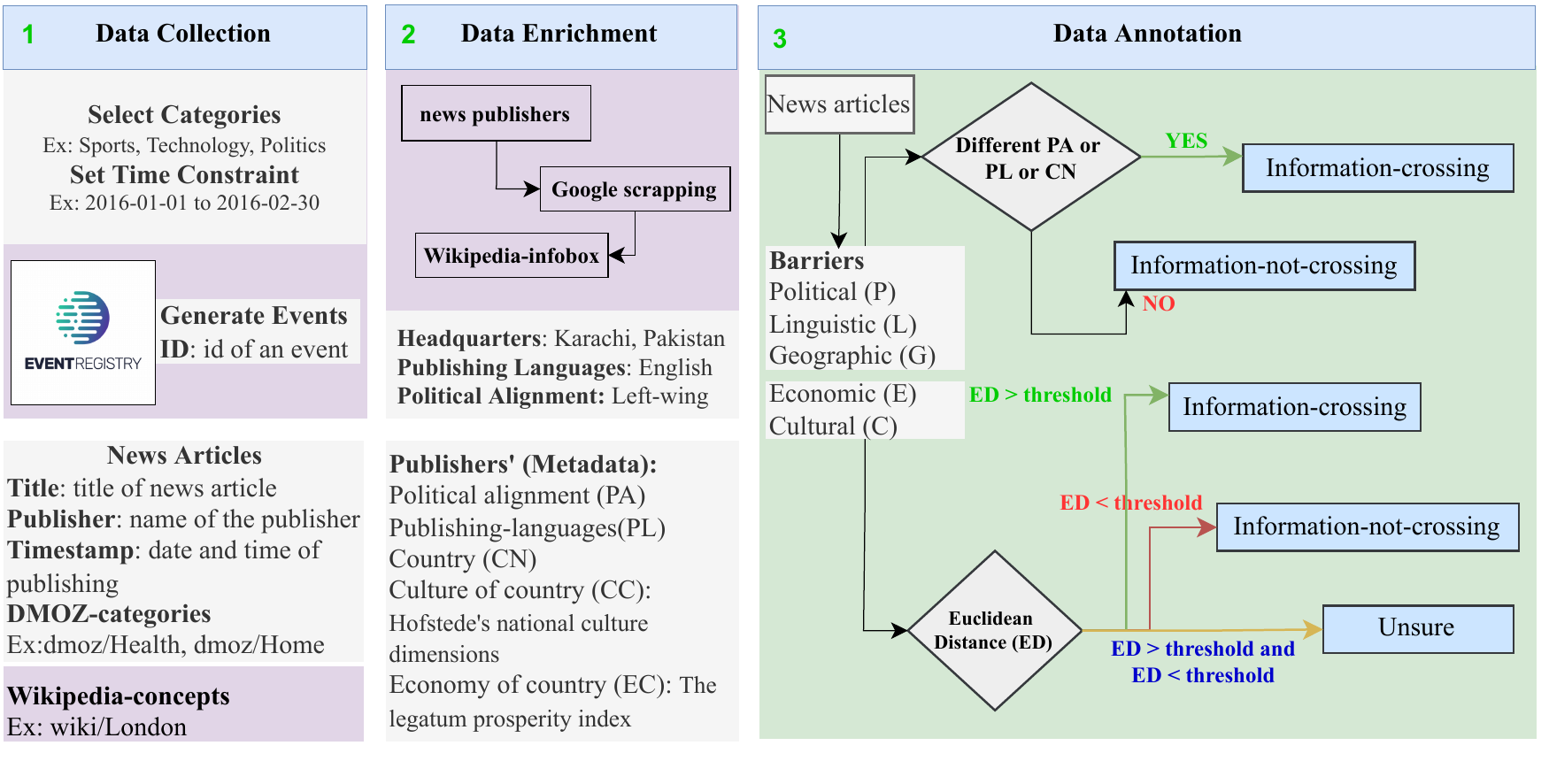}
\caption{\bf An approach to barrier classification based on news meta-data. Data extraction from the Event Registry is the first step. Meta-data extraction through Google and Wikipedia scrapping is the second step. The third step is to annotate the news articles after calculating the euclidean distances.}
\label{fig:approach}
\end{figure*}

The presented research focuses on barrier classification in news articles. To this end, we propose a novel approach to barrier classification based on news meta-data, as shown in Figure \ref{fig:approach}.
\newline
In the first step, we execute a query that extracts the news articles from the Event Registry belonging to different categories (business, computers, games, health, home, recreation, science, shopping, society, and sports) and published within a certain time span - in our case between 2016-2021 (see Subsection \ref{sec:datadescription}). Then we parse and save these news articles along with the source information such as the publishers' names and publishing dates. In the second step, we extract the meta-data related to the news publishers via searching the news publishers' on Google and extracting their Wikipedia links. Using this link, we obtain the necessary information from Wikipedia-infobox (see Subsection \ref{subsec:metadataexpl}). In the third step, we perform the annotation of news articles. To annotate the news articles, we set the annotation guidelines \ref{subsec:metadataexpl}. For cultural and economic barriers, we assign the ternary labels to news articles whereas, for the linguistic, geographical, and political barriers, we assign the binary labels to the news articles. Table \ref{tab:annotationExamples} presents the examples of annotation for all the barriers. Afterward, we conduct experiments comparing machine learning state-of-the-art classification methods, deep learning, and transformer-based methods (see Figure \ref{fig:overview}). The results are presented in Section \ref{subsec:tencats}, \ref{subsec:algos} showing the performance of different features and different methods. 

\begin{figure}
\centering
\includegraphics[keepaspectratio=true,scale=0.30]{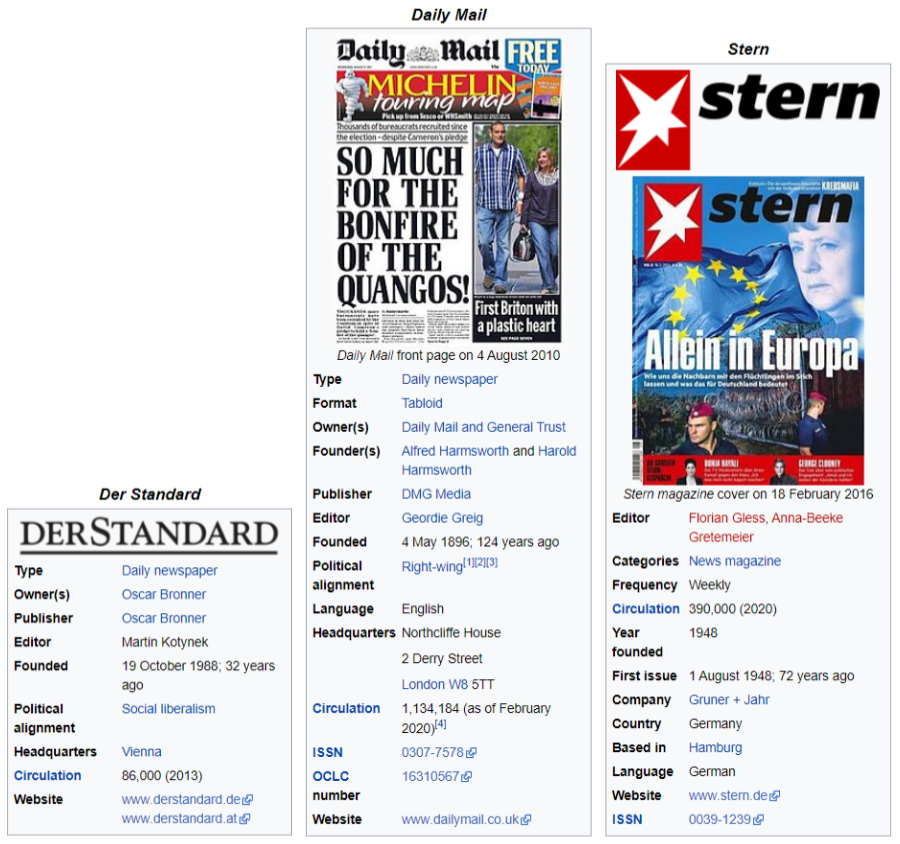}
\caption{\bf Three Wikipedia-infobox for the three different newspapers/magazines with their political alignment}
\label{fig:infobox}
\end{figure}

\section{Dataset description}\label{sec:datadescription}
We collected the news articles reporting on different events published between 2016-2021 in the English language using Event Registry \cite{ref:leban2014event} APIs \footnote{\url{https://github.com/EventRegistry/event-registry-python/blob/master/eventregistry/examples/QueryArticlesExamples.py}}. The dataset consists of 35 million news articles that take storage up to 150 GB. Each news article belongs to a different category (see Figure \ref{fig:totalfields}). Each news article consists of a few attributes: title, body text, name of the news publisher, date and time of publishing, event-ID, DMOZ-categories, and Wikipedia concepts. 

\begin{figure}
\centering
\includegraphics[keepaspectratio=true,scale=0.35]{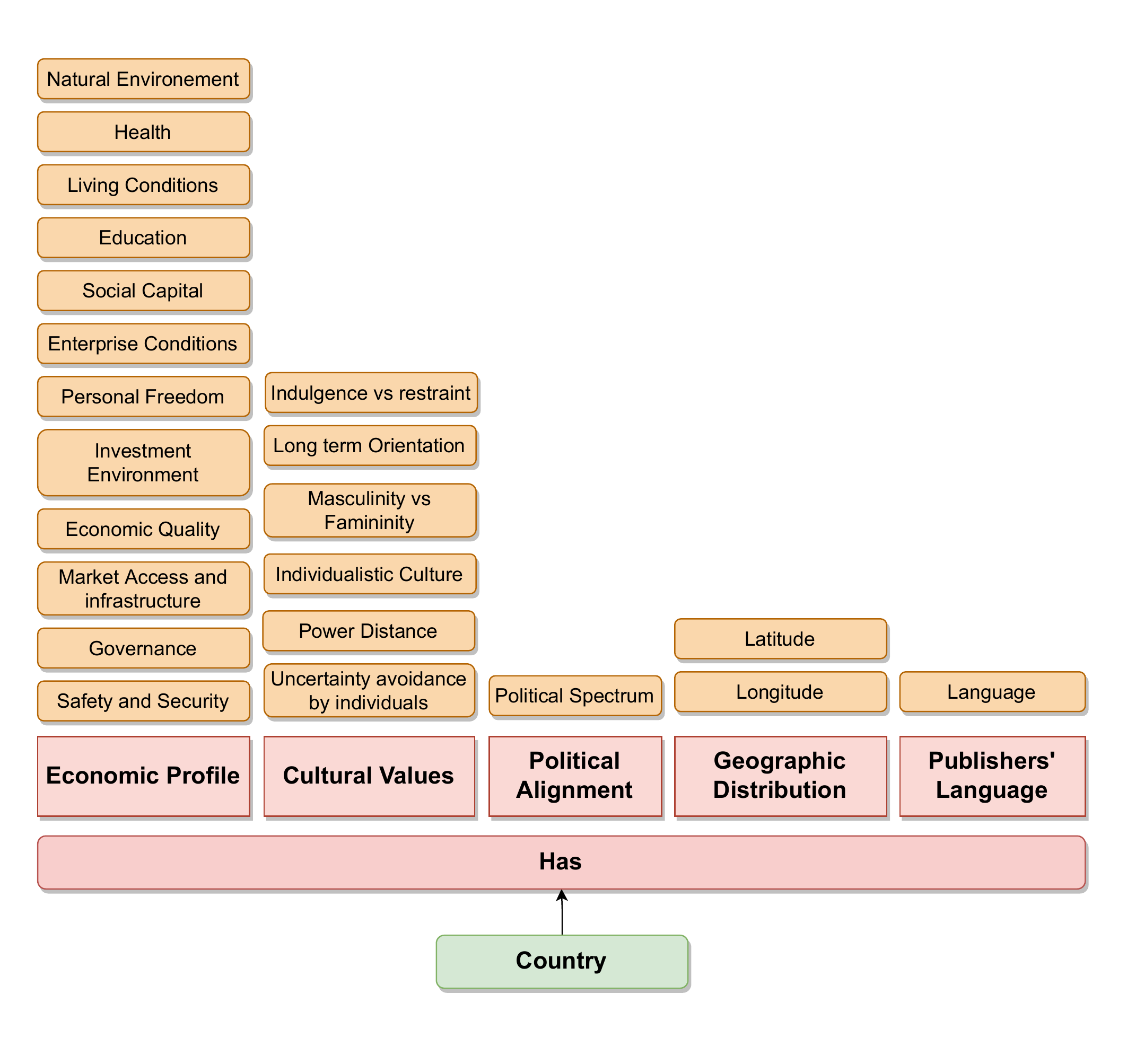}
\caption{Metadata for the five barriers (cultural, economic, geographical, linguistic, and political)}
\label{fig:schema}
\end{figure}

\begin{figure*}
\centering
        \includegraphics[width=0.32\textwidth]{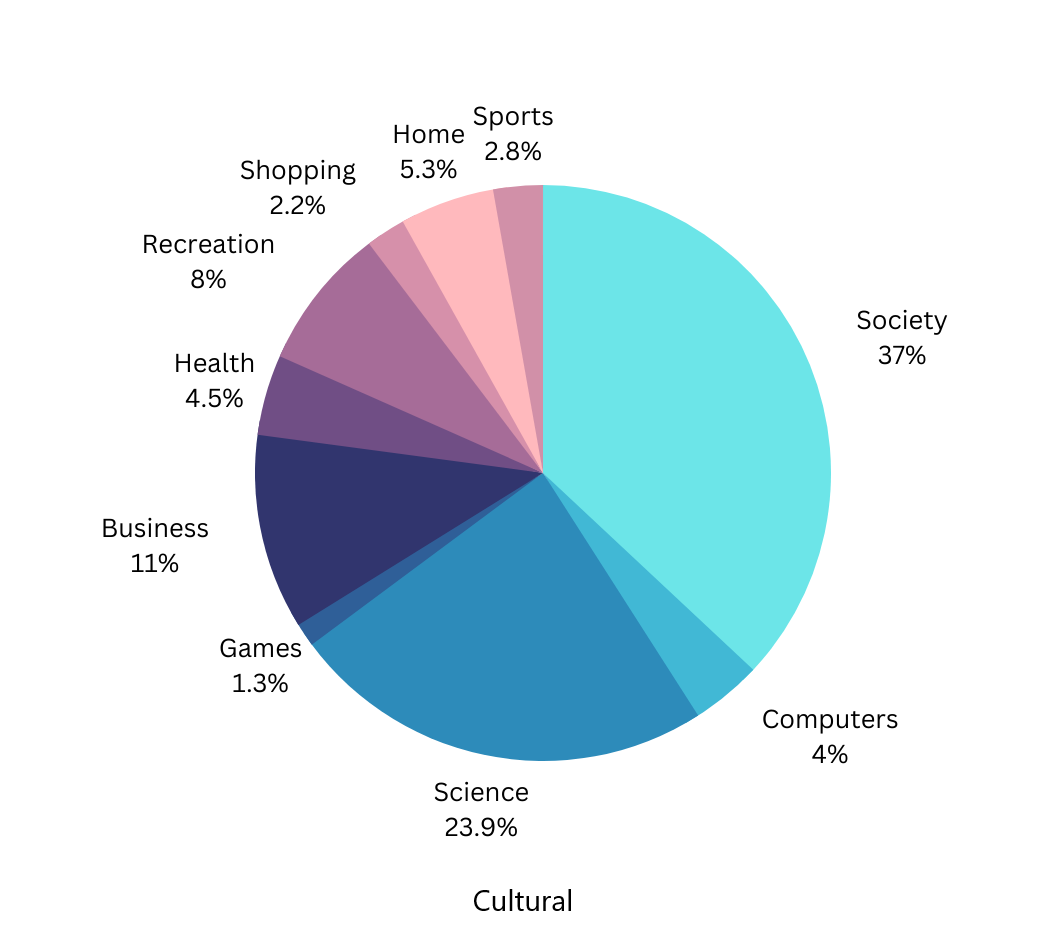}
        \includegraphics[width=0.32\textwidth]{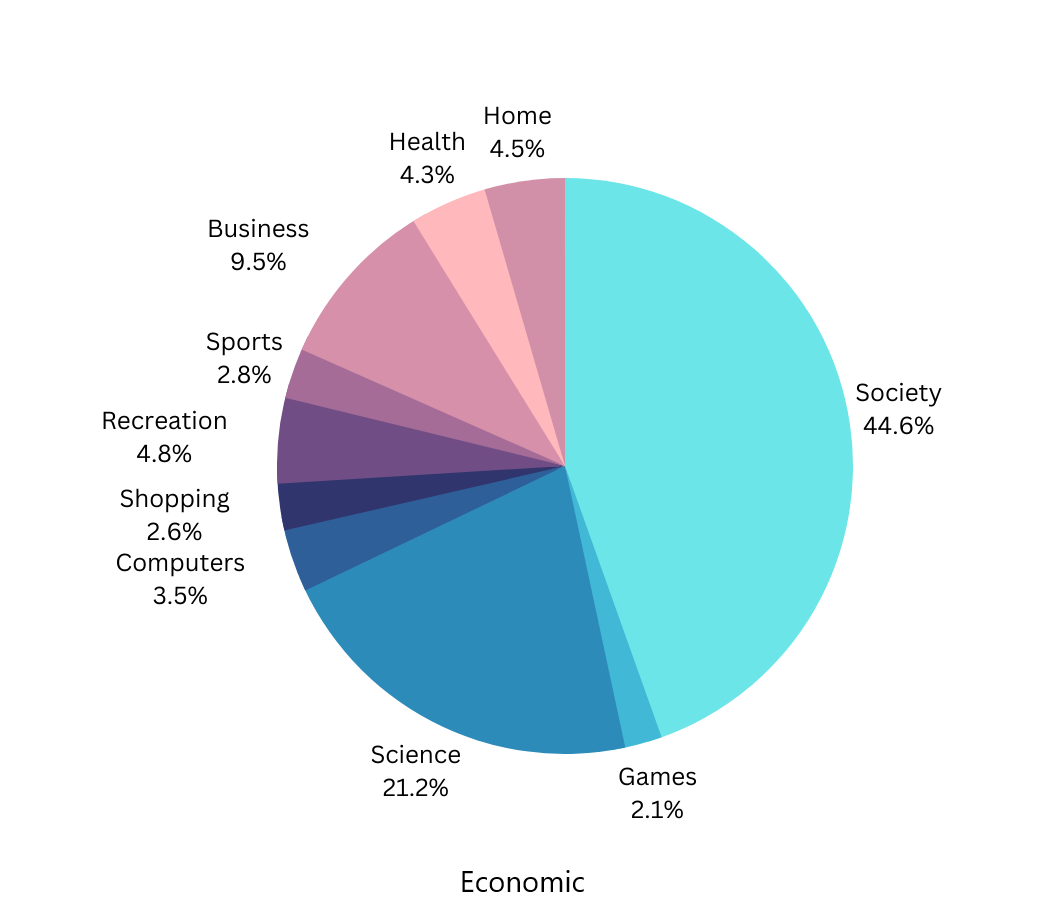}
        \includegraphics[width=0.32\textwidth]{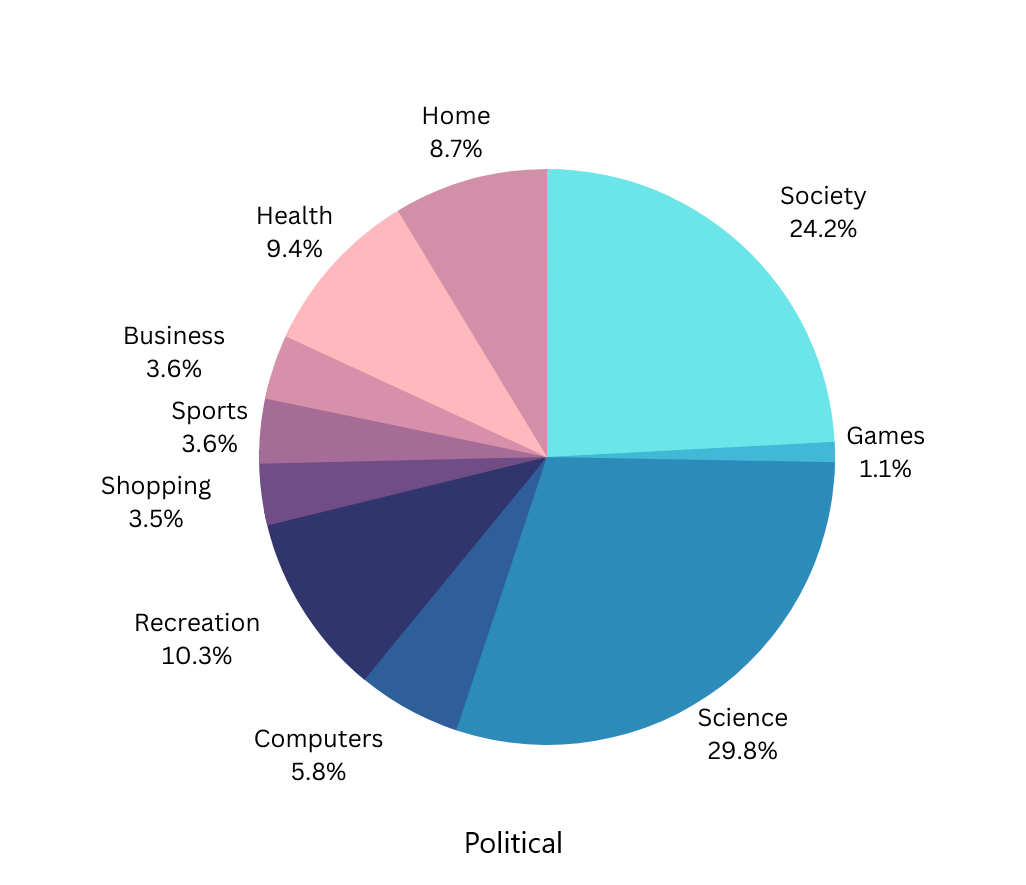}
        \includegraphics[width=0.32\textwidth]{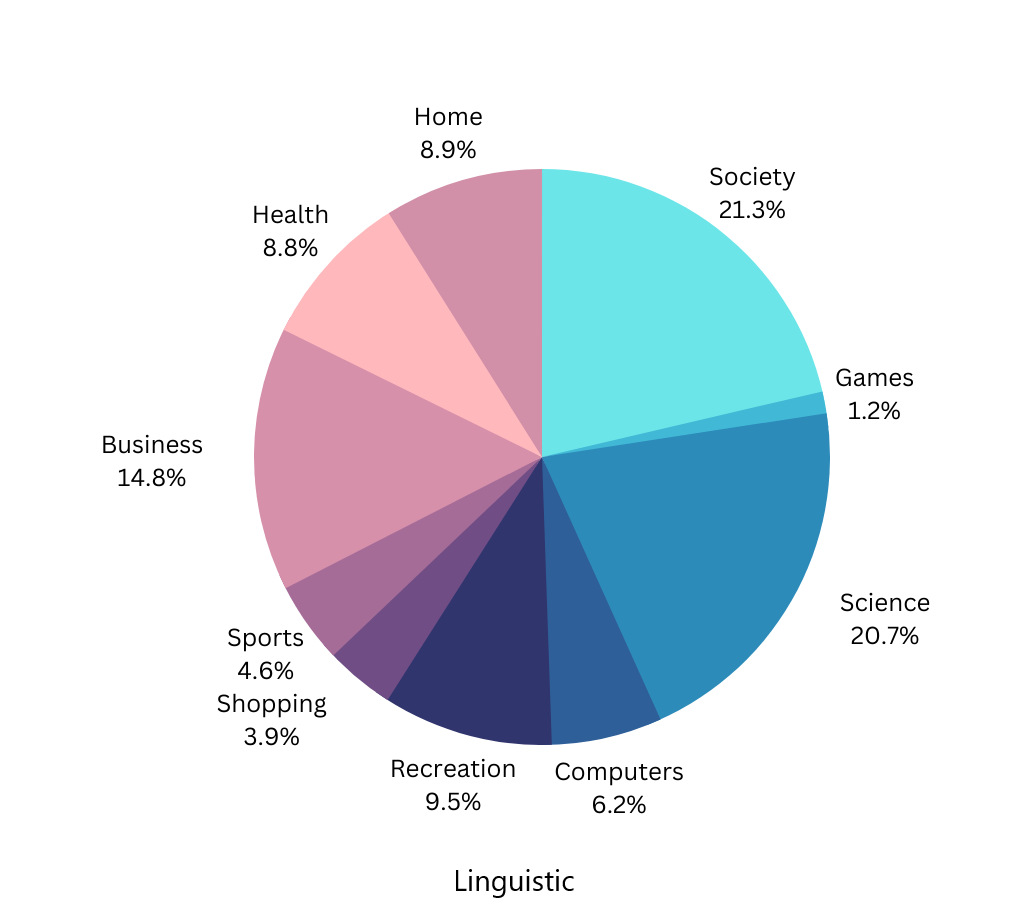}
        \includegraphics[width=0.32\textwidth]{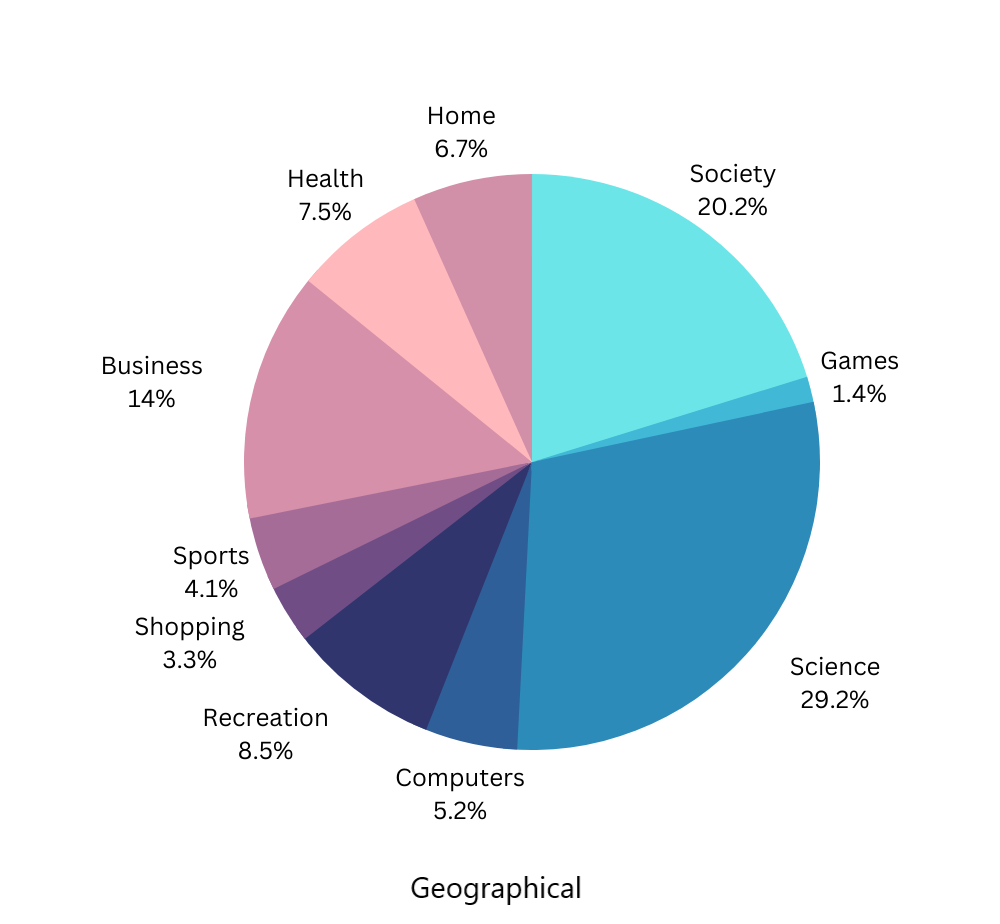}
        \caption{\bf The pie charts show the statistics about the news articles for the five news spreading barriers (from left to right: cultural, economic, political, linguistic, and geographic) that belong to ten different categories (business, computers, games, health, home, recreation, science, shopping, society, and sports). We can see that a more percentage of news articles belong to science, society, and business categories.}
        \label{fig:catsstats}
\end{figure*}

A few attributes are self-explanatory such as title, body text, name of the news publisher, and date and time of publishing. An event-id represents a unique number that is associated with all the news articles that belong to a same event. The DMOZ-categories represent the topics of the content/news article. It is a project that has hierarchical collection of web page links organized by subject matters \footnote{\href{https://dmoz-odp.org/}{https://dmoz-odp.org/}}. Around 50,000 categories are used by the Event Registry (top 3 layers of the DMoz taxonomy)  \footnote{\href{https://eventregistry.org/documentation?tab=terminology}{https://eventregistry.org/documentation?tab=terminology}}. 
The statistics of all the categories for all the five barriers are presented in the pie charts (see Figure \ref{fig:catsstats}). Wikipedia concepts are used as a semantic annotation for the news articles and can represent entities (locations, people, organizations) or non-entities (things such as personal computers, and toys). In Event Registry, Wikipedia's URLs are used as concept URIs. 

\subsection{Similarity between news articles}\label{subsec:similarity}
Event Registry is a platform that collects multi-lingual similar news articles from tens of thousands of news sources and identifies events\cite{ref:leban2014event}. It collects data using the News Feed service \cite{ref:trampuvs2012internals} which collects news articles from around 75.000 news sources in various languages (English, German, Spanish, and Chinese). To construct an event, it groups similar news articles. It calculates many features, and cross-lingual similarity of articles is one of them. It does not use any machine translators, but rather tries to frame the problem of finding similarities among cross-lingual news articles such as that they could use well-established machine learning tools designed for mono-lingual text-mining tasks. It looks at the distribution of articles across languages where English was the largest language and use as one of the hub languages which not only has an order of magnitude with more articles than other languages, but also many comparable articles with most of the other languages. 

\subsection{Metadata for each barrier}\label{subsec:metadataexpl}
To fetch the metadata for each barrier, the essential thing is the news publisher's headquarters name. For each news publisher we get this information from  Wikipedia-infobox (see Figure \ref{fig:infobox}). We used Bright Data service \footnote{\url{https://brightdata.com/}} to crawl and parse Wikipedia-Infobox for almost more than 10,000 news websites. We retrieved the country name of the news publisher's headquarters name. For the economical barrier, we fetched the economical profile for each country using "The Legatum Prosperity Index" \footnote{\url{https://www.prosperity.com/}} as done by \cite{ref:sittar2022analysis}. It has twelve dimensions that represent different economical aspects (see Figure \ref{fig:schema}). For the cultural barrier, we calculated differences among different regions using six Hofstede’s national culture dimensions (HNCD) (see Figure \ref{fig:schema}). For the economic and cultural barrier, we calculated the euclidean distance among all the countries (for the economic barrier using the economical profile, and for the cultural barrier using the HNCD). Two countries have been labelled as: "information-not-crossing" if the distance score was $\leq$ 0.1, "unsure" if the distance score was $>$ 0.1 and $\leq$ 0.4 , "information-crossing" if the distance score was $>$ 0.4 (see examples in the Table \ref{tab:annotationExamples}). 

\begin{figure*}
            \centering
            \includegraphics[keepaspectratio=true,scale=0.18]{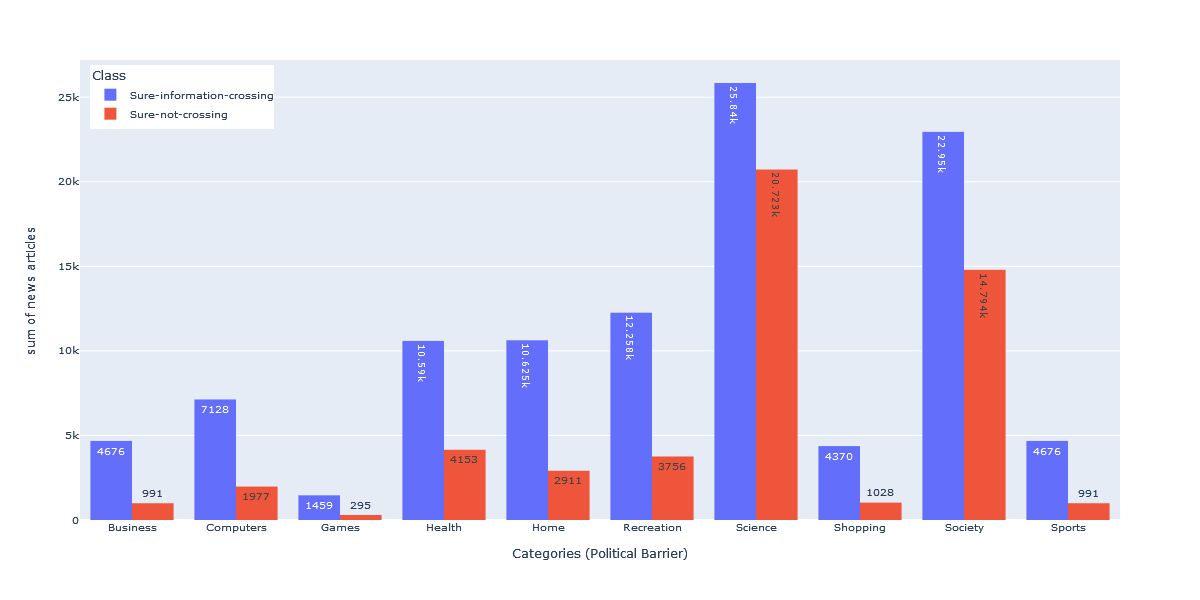}
            \includegraphics[keepaspectratio=true,scale=0.18]{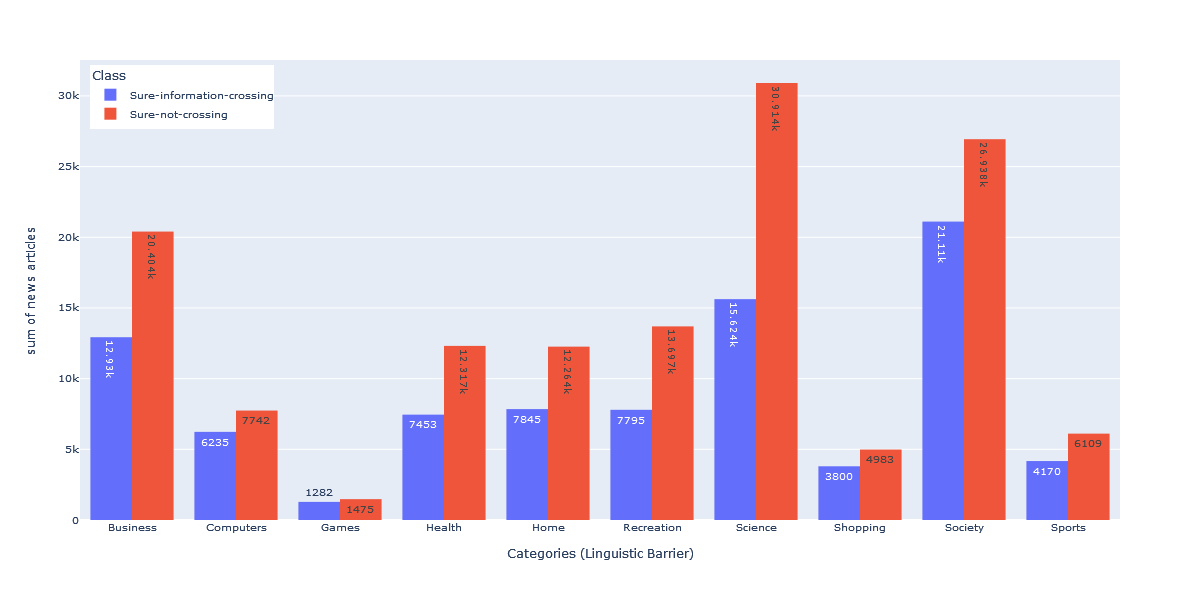}
            \includegraphics[keepaspectratio=true,scale=0.36]{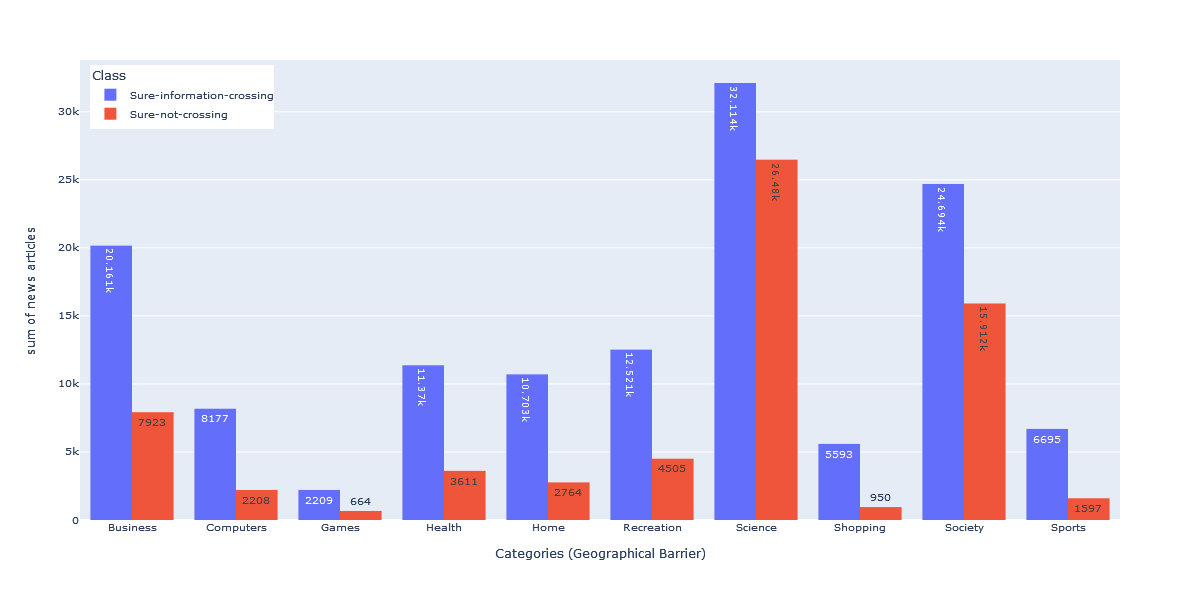}
            \caption{\bf This bar chart shows the class distribution for the political, linguistic, and geographic barriers (from left to right). The bar with blue color shows the distribution for the class "Information-crossing" a barrier whereas the bar with red color shows the distribution for the class "Information-not-crossing" a barrier. Each of the three-bar charts presents the class distribution for all the ten categories.}
            \label{fig:barriDistclass}
\end{figure*}

For the geographical barrier, we stored general latitude and longitude. For the political barrier, we utilize the political ideology/alignment of the newspaper/magazine that we determined based on Wikipedia-infobox at their Wikipedia page \cite{ref:sittar2022political}(see Figure \ref{fig:infobox}). The statistics about the annotated dataset are presented in Figure \ref{fig:barriDistclass}, and \ref{fig:ecoculDist}. The data is proprietary to Event Registry \footnote{\href{https://eventregistry.org/}{https://eventregistry.org/}}. People can ask if they need that kind of data. 

\begin{table*}
\centering
\caption{\bf This table shows the examples of annotation for all the five types of barriers. The annotation is performed using the meta-data shown in the Figure \ref{fig:schema}.}
\scalebox{0.5}{
\label{tab:annotationExamples}
\begin{tabular}{|l|l|l|l|l|l|} 
\hline
\textbf{Barrier}     & \textbf{Time}                                                                                                                                   & \textbf{Title}                                                                                                                                                                                                                                                                                                                                                                                & \begin{tabular}[c]{@{}l@{}}\textbf{Location/Publisher/}\\\textbf{Language}\end{tabular}              & \textbf{Meta-Data}                                                                        & \textbf{Class}                                                       \\ 
\hline
\textbf{Cultural }   & \begin{tabular}[c]{@{}l@{}}2019-01-11 00:17:00\\2019-01-11 07:54:00\\2019-01-11 17:14:00\\2019-01-11 18:57:00\\2019-01-11 20:52:00\end{tabular} & \begin{tabular}[c]{@{}l@{}}Timeline: Life and death of ex-Nazi guard deported from US \textbar{} The Star\\Former Nazi camp guard deported by US dies in Germany\\Former Nazi guard deported after years of living in U.S. dies at 95\\Former Nazi guard who spent decades living in U.S. before deportation dies\\Former Nazi Guard Dies After Being Deported From US In August\end{tabular} & \begin{tabular}[c]{@{}l@{}}U.S.A\\Canada\\Switzerland\\France\\Israel\end{tabular}                   & Same Culture                                                                              & information-not-crossing                                                    \\ 
\cline{2-6}
                     & \begin{tabular}[c]{@{}l@{}}2019-01-11 07:53:00\\2019-01-11 11:49:00\\2019-01-11 17:24:00\end{tabular}                                           & \begin{tabular}[c]{@{}l@{}}Trump cancels planned Davos trip as shutdown drags on\\Fact check: will Mexico pay for Trump's border wall?\\Congress is taking the weekend off as federal workers face their first empty payday\end{tabular}                                                                                                                                                      & \begin{tabular}[c]{@{}l@{}}U.S.A\\Germany\\U.S.A\end{tabular}                                        & Different Culture                                                                         & un-sure                                                              \\ 
\cline{2-6}
                     & \begin{tabular}[c]{@{}l@{}}2019-01-16 10:38:00\\2019-01-17 12:09:00\\2019-01-18 06:34:00\\2019-01-18 09:54:00\end{tabular}                      & \begin{tabular}[c]{@{}l@{}}LUNGU DATES ETHIOPIA\\African leaders to meet over DRC vote dispute\\Amid calls for govt of national unity, AU urges delay in announcing~\\DRC election results \textbar{} Ripples Nigeria\\African Union urges DRC to delay final election results\end{tabular}                                                                                                   & \begin{tabular}[c]{@{}l@{}}South Africa\\Zambia\\Nigeria\\Turkey\end{tabular}                        & Different Culture                                                                         & \begin{tabular}[c]{@{}l@{}}sure-information-\\crossing\end{tabular}  \\ 
\hline
\multicolumn{6}{|l|}{}                                                                                                                                                                                                                                                                                                                                                                                                                                                                                                                                                                                                                                                                                                                                                                                                                           \\ 
\hline
\textbf{Economic }   & \begin{tabular}[c]{@{}l@{}}2016-12-24 16:10:00\\2016-12-24 22:01:00\\2016-12-25 03:51:00\end{tabular}                                           & \begin{tabular}[c]{@{}l@{}}Thousands of faithful celebrate Christmas in Bethlehem\\Believers gather in Bethlehem for Christmas at birthplace of Jesus\\Faithful celebrate Christmas in Bethlehem\end{tabular}                                                                                                                                                                                 & \begin{tabular}[c]{@{}l@{}}Canada\\Ireland\\U.S.A\end{tabular}                                       & \begin{tabular}[c]{@{}l@{}}Similar Economic \\Situations (ES)\end{tabular}                & information-not-crossing                                                    \\ 
\cline{2-6}
                     & \begin{tabular}[c]{@{}l@{}}2016-12-28 17:07:00\\2016-12-28 23:48:00\end{tabular}                                                                & \begin{tabular}[c]{@{}l@{}}Germany detains Tunisian for possible link to Berlin attack\\Berlin attacker thought to have fled through Netherlands \textbar{} MACAU DAILY TIMES\end{tabular}                                                                                                                                                                                                    & \begin{tabular}[c]{@{}l@{}}Oman\\China\end{tabular}                                                  & Different ES                                                                              & un-sure                                                              \\ 
\cline{2-6}
                     & \begin{tabular}[c]{@{}l@{}}2016-12-25 05:34:00\\2016-12-25 08:39:00\\2016-12-26 01:25:00\\2016-12-26 05:07:00\end{tabular}                      & \begin{tabular}[c]{@{}l@{}}Donald Trump says he will dissolve foundation amid New York investigation\\Trump to shut down his charitable foundation to avoid conflict of interests\\Trump says he intends to dissolve charitable foundation 26-Dec-16 169\\To conceal a conflict of interest, Donald Trump dissolves foundation - News\end{tabular}                                            & \begin{tabular}[c]{@{}l@{}}India\\Nigeria\\Pakistan\\India\end{tabular}                              & Different ES                                                                              & \begin{tabular}[c]{@{}l@{}}sure-information-\\crossing\end{tabular}  \\ 
\hline
                     &                                                                                                                                                 &                                                                                                                                                                                                                                                                                                                                                                                               &                                                                                                      &                                                                                           &                                                                      \\ 
\hline
\textbf{Political }  & \begin{tabular}[c]{@{}l@{}}2016-01-27 15:48:00\\2016-01-27 20:15:00\\2016-01-29 04:18:00\\2016-01-29 18:23:00\end{tabular}                      & \begin{tabular}[c]{@{}l@{}}Clinton wants to do 'unsanctioned' debate but Bernie won't commit\\Bernie Sanders says no to unsanctioned debate\\Sanders, Clinton, O'Malley on board for NH debate, await word from DNC\\Democratic debate plans in New Hampshire are still in flux\end{tabular}                                                                                                  & \begin{tabular}[c]{@{}l@{}}dailymail.co.uk\\usatoday.com\\unionleader.com\\usatoday.com\end{tabular} & \begin{tabular}[c]{@{}l@{}}Similar Political \\alignment (PA)\end{tabular}                & information-not-crossing                                                    \\ 
\cline{2-6}
                     & \begin{tabular}[c]{@{}l@{}}2016-02-16 22:00:00\\2016-02-16 22:13:00\\2016-02-17 02:27:00\end{tabular}                                           & \begin{tabular}[c]{@{}l@{}}B.C. budget 2016: Fourth consecutive balanced budget sees tweaks on housing~\\taxes, MSP\\B.C.'s balanced budget aims to cool province's red-hot real estate market\\B.C. announces funding boost for children in government care\end{tabular}                                                                                                                     & \begin{tabular}[c]{@{}l@{}}ottawacitizen.com\\theglobeandmail.com\\theglobeandmail.com\end{tabular}  & Different PA                                                                              & \begin{tabular}[c]{@{}l@{}}sure-information-\\crossing\end{tabular}  \\ 
\hline
\multicolumn{6}{|l|}{}                                                                                                                                                                                                                                                                                                                                                                                                                                                                                                                                                                                                                                                                                                                                                                                                                           \\ 
\hline
\textbf{Linguistic } & \begin{tabular}[c]{@{}l@{}}2019-01-01 00:09:00\\2019-01-01 00:55:00\\2019-01-01 01:07:00\end{tabular}                                           & \begin{tabular}[c]{@{}l@{}}Warren's jump into the presidential campaign kicks the 2020 race into high gear\\Democrat Warren enters 2020 White House race\\Warren's jump into the presidential campaign kicks the 2020 race into high gear\end{tabular}                                                                                                                                        & \begin{tabular}[c]{@{}l@{}}English\\English\\English\end{tabular}                                    & \begin{tabular}[c]{@{}l@{}}Similar Publishing \\Language (PL)\end{tabular}                & information-not-crossing                                                    \\ 
\cline{2-6}
                     & \begin{tabular}[c]{@{}l@{}}2019-01-29 08:33:00\\2019-01-29 11:48:00\\2019-01-29 15:10:00\end{tabular}                                           & \begin{tabular}[c]{@{}l@{}}Switzerland ranked among least corrupt in global index\\SA was more corrupt in 2018 than it was in 2017, survey shows\\Pakistan one point up in Corruption Perceptions Index 2018\end{tabular}                                                                                                                                                                     & \begin{tabular}[c]{@{}l@{}}Portuguese\\English\\English\end{tabular}                                 & Different PL                                                                              & \begin{tabular}[c]{@{}l@{}}sure-information-\\crossing\end{tabular}  \\ 
\hline
\multicolumn{6}{|l|}{}                                                                                                                                                                                                                                                                                                                                                                                                                                                                                                                                                                                                                                                                                                                                                                                                                           \\ 
\hline
\textbf{Geographic } & \begin{tabular}[c]{@{}l@{}}2018-02-01 20:53:00\\2018-02-02 11:26:00\\2018-02-02 01:50:00\end{tabular}                                           & \begin{tabular}[c]{@{}l@{}}The Latest: Polish PM: Poles, Jews share need for WWII truth\\Poland's Holocaust controversy and Donald Trump's rubber-stamp of Polish~\\nationalism\\Polish PM: We understand Israel's emotions\end{tabular}                                                                                                                                                      & \begin{tabular}[c]{@{}l@{}}U.S.A\\Israel\\Israel\end{tabular}                                        & \begin{tabular}[c]{@{}l@{}}Publishers' \\headquarters in \\same country\end{tabular}      & information-not-crossing                                                    \\ 
\cline{2-6}
                     & \begin{tabular}[c]{@{}l@{}}2018-05-16 01:46:00\\2018-05-16 02:11:00\\2018-05-16 02:57:00\end{tabular}                                           & \begin{tabular}[c]{@{}l@{}}The Latest: Pete Ricketts wins GOP Nebraska governor primary\\The Latest: Bob Krist wins Nebraska Democratic governor nod\\Voters select Ricketts, Krist in gubernatorial primary\end{tabular}                                                                                                                                                                     & \begin{tabular}[c]{@{}l@{}}U.S.A\\U.S.A\\U.S.A\end{tabular}                                          & \begin{tabular}[c]{@{}l@{}}Publishers' \\headquarters in \\different country\end{tabular} & \begin{tabular}[c]{@{}l@{}}sure-information-\\crossing\end{tabular}  \\ 
\hline
\multicolumn{1}{l}{} & \multicolumn{1}{l}{}                                                                                                                            & \multicolumn{1}{l}{}                                                                                                                                                                                                                                                                                                                                                                          & \multicolumn{1}{l}{}                                                                                 & \multicolumn{1}{l}{}                                                                      & \multicolumn{1}{l}{}                                                
\end{tabular}}
\end{table*}

\begin{figure*}
\centering
        \includegraphics[width=0.45\textwidth]{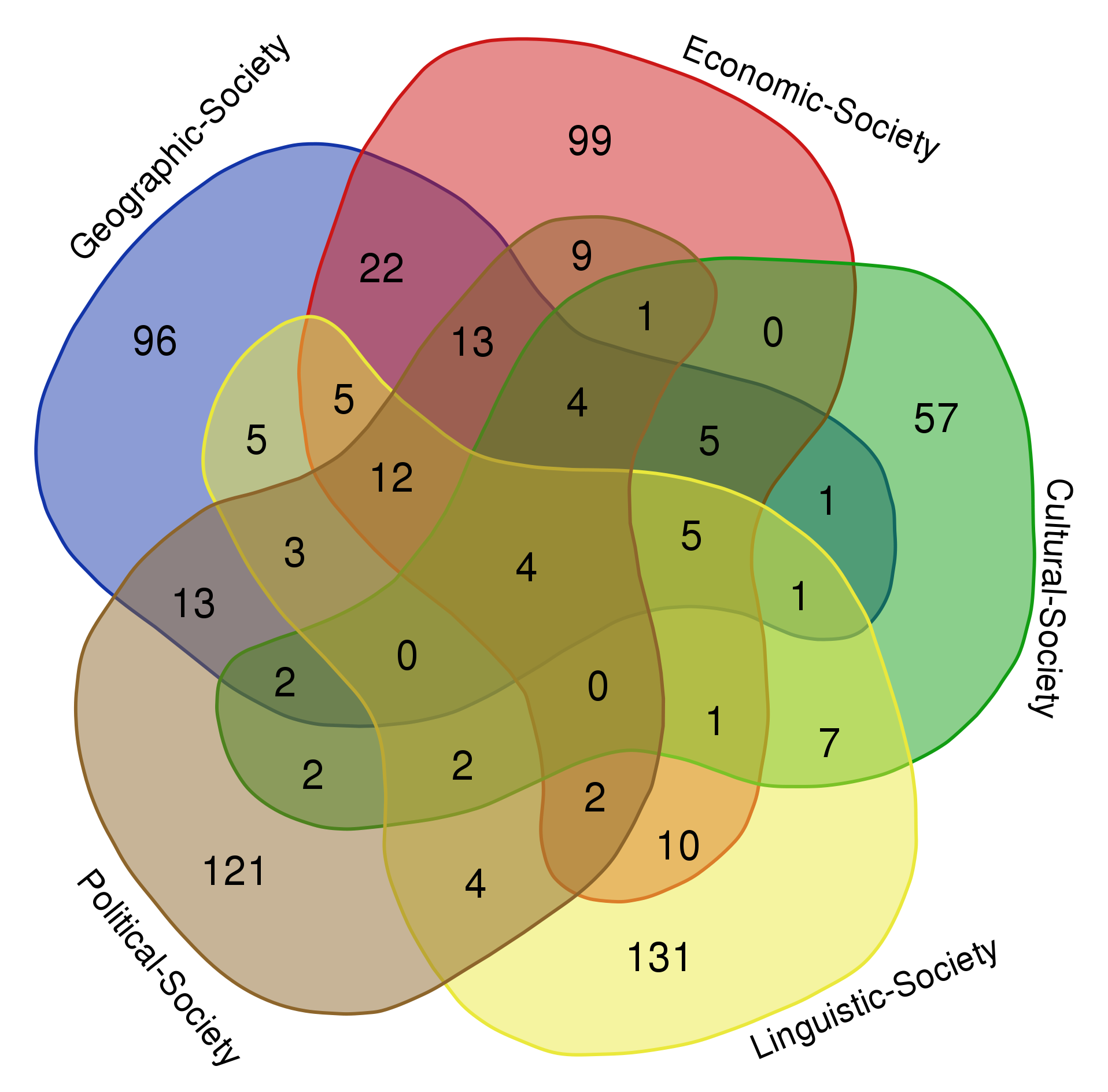}
        \hspace{2cm}
        \includegraphics[width=0.40\textwidth]{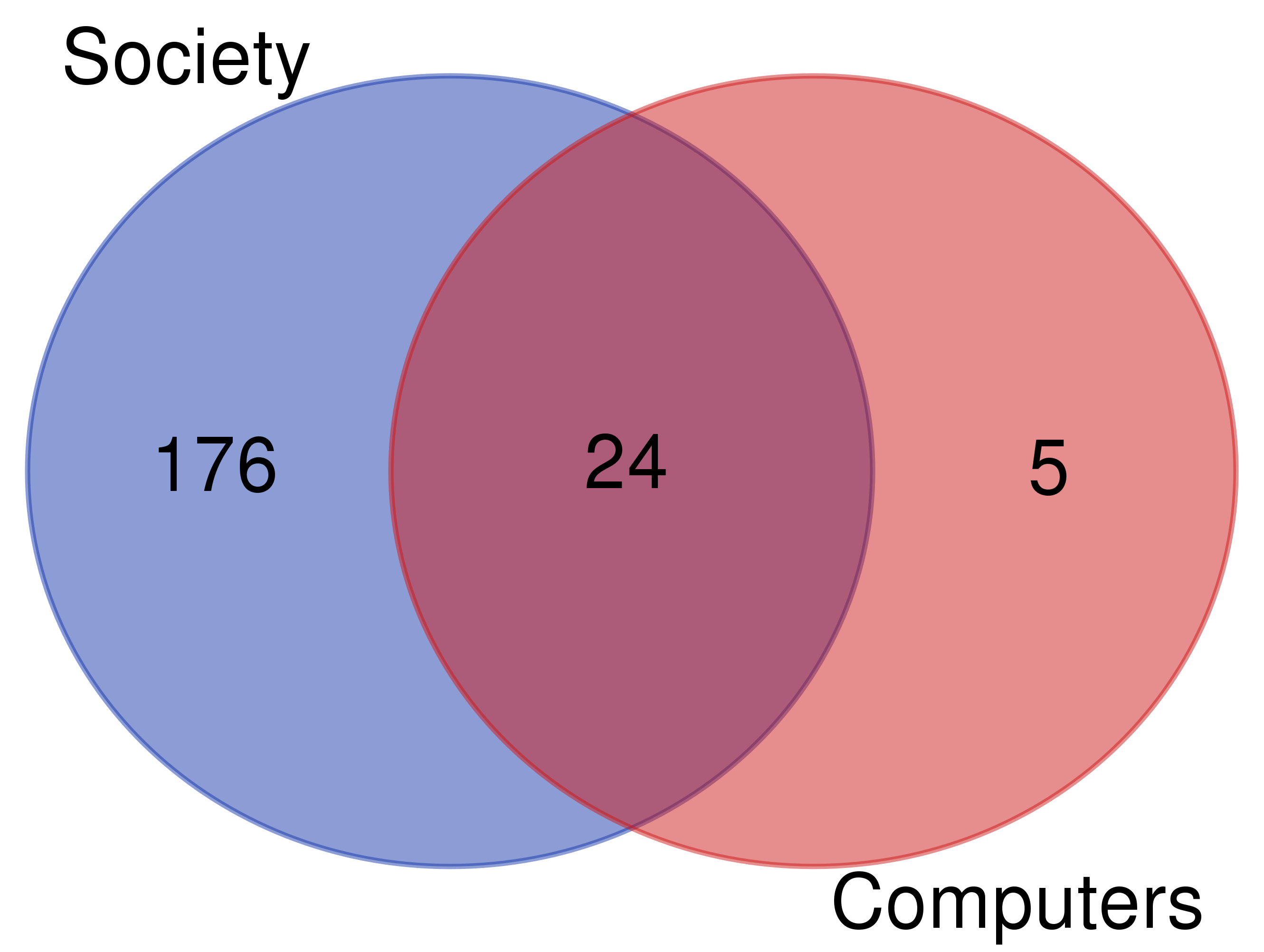}
        \caption{\bf The Venn diagrams show the intersection between the Wikipedia concepts across the five barriers and different categories.}
        \label{fig:venndiagrams}
\end{figure*}

\textbf{Annotation Questions}:
Based on the definitions above, we set the following annotation questions in order to identify barriers to news spreading.
\begin{itemize}
\setlength\itemsep{0em}
\item Q1: \emph{Do all the news articles reporting on an event, publish from a particular/same geographical location?}
\item Q2: \emph{Do all the news articles reporting on an event, publish from the locations having equal economic prosperity?}
\item Q3: \emph{Do all the news articles reporting on an event, publish from a particular/same locations having equal cultures?}

\item Q4: \emph{Do all the news articles reporting on an event, publish from the sources with a particular/similar political class?}

\item Q5: \emph{Do all the news articles reporting on an event, publish by the newspapers where the publishing language were same?}
\end{itemize}

Question 1~(Q1) intends to identify whether the news was published across different geographical places or not. The question is answered "Yes" for all the news articles reported on an event if they are published from one country otherwise "No". Question 2~(Q2) intends to identify whether the news was published across different economies or not. The economic similarity has been calculated using euclidean distance. The question is answered with "information-crossing" for all the news articles reported on an event if they are published from countries with similar economic situations. The question is answered with "unsure" for all the news articles reported on an event if at least one of the news articles published from a country that is labeled with "unsure" (see Subsection \ref{subsec:metadataexpl}) otherwise "information-not-crossing". Question 3~(Q3) intends to identify whether the news was published across different cultures or not. The question is answered with "information-crossing" for all the news articles reported on an event if they are published from countries with a similar culture. The question is answered with "unsure" for all the news articles reported on an event if at least one of the news articles is published from a country that is labeled with "unsure" otherwise "information-not-crossing". The cultural similarity has been calculated using euclidean distance (see Subsection \ref{subsec:metadataexpl}). Question 4~(Q4) intends to identify whether the news was published in newspapers with the same political alignments or not. The question is answered "Yes" for all the news articles reporting on an event if they are published in the newspapers following similar political alignments otherwise "No". Question 5~(Q5) intends to identify whether the news was published in the newspapers where the publishing language was the same or not. The question is answered "Yes" for all the news articles reporting on an event if they are published from different newspapers where the publishing language was same otherwise "No".

\subsubsection{Barrier Categories}\label{subsubsec:dataset_labels}
Labels for the five types of barrier annotations are derived:
\begin{itemize}
\setlength\itemsep{0em}
\item Economic barrier classes: \textit{information-not-crossing}, \textit{unsure}, and \textit{information-crossing}.
\item Cultural barrier classes: \textit{information-not-crossing}, \textit{unsure}, and \textit{information-crossing}.
\item Geographical barrier classes: \textit{Not-crossed-GB}, and \textit{Crossed-GB}.
\item Political barrier classes: \textit{Not-crossed-PB}, and \textit{Crossed-PB}.
\item Linguistic barrier classes: \textit{Not-crossed-LB}, and \textit{Crossed-LB}.
\end{itemize}

\subsubsection{Analysis of information spreading and Wikipedia concepts}
\textbf{Q1: Does the information spreading in news varies across different topics and different barriers?}

\begin{figure*}
            \centering
            \includegraphics[keepaspectratio=true,scale=0.29]{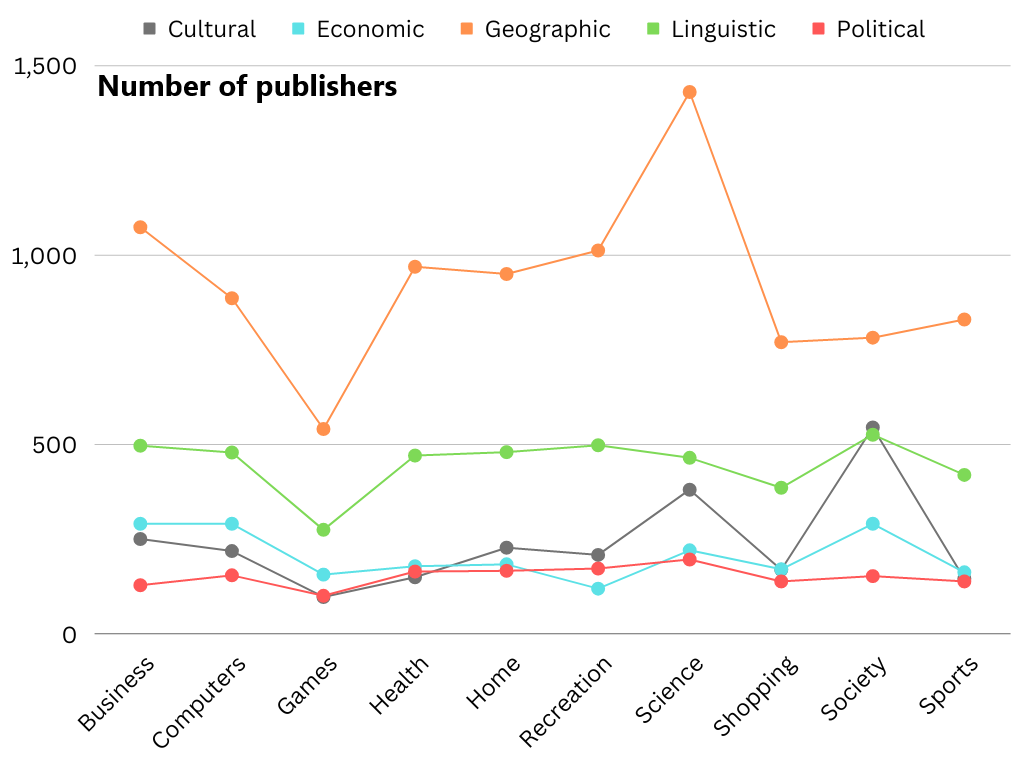}
            \includegraphics[keepaspectratio=true,scale=0.29]{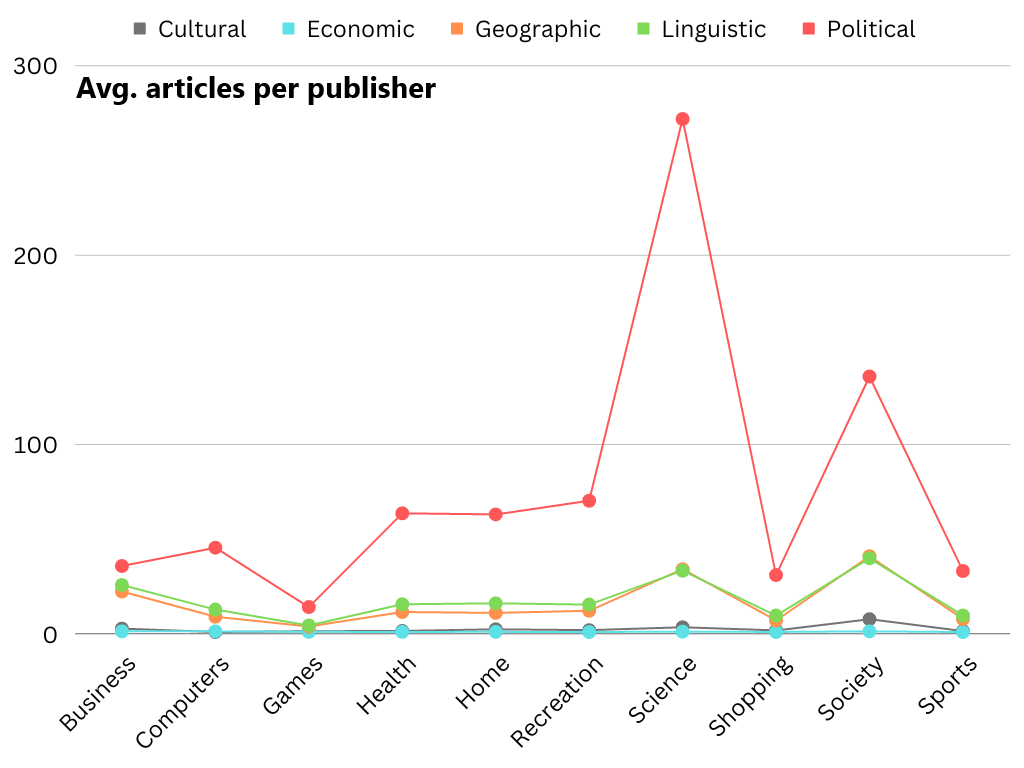}
            \includegraphics[keepaspectratio=true,scale=0.56]{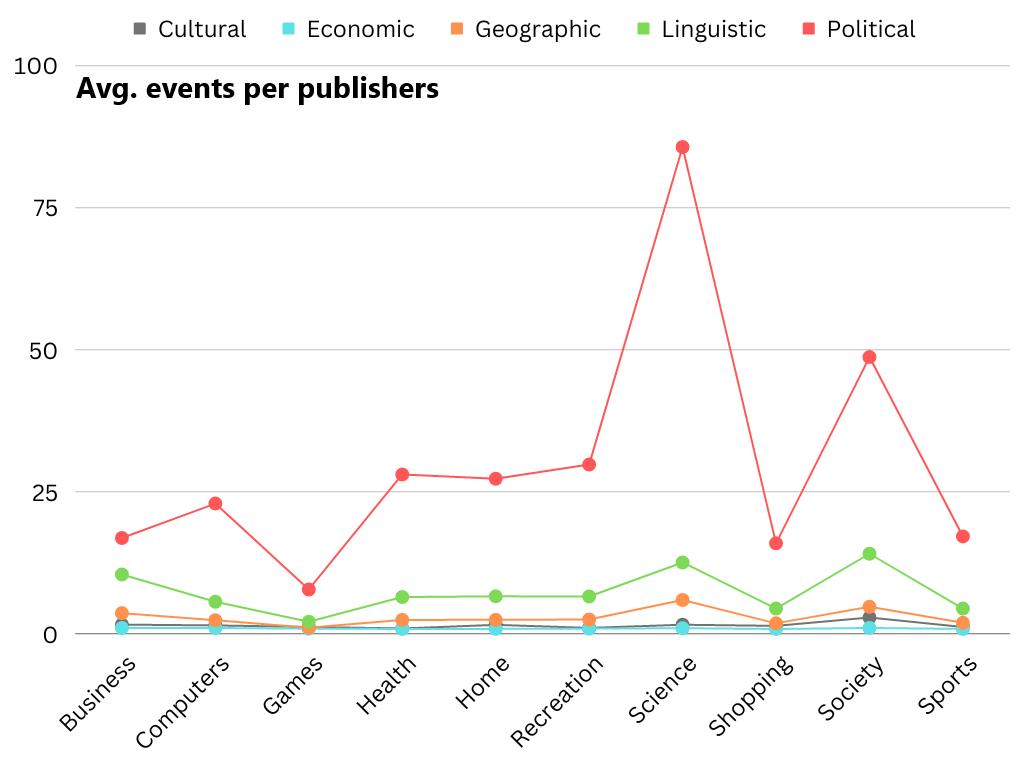}
            \caption{\bf These line charts show the number of publishers, the average number of news articles per publisher, and the average number of events per publisher (from left to right). The lines with red, green, orange, blue, and gray colors represent the political, linguistic, geographic, economic, and cultural barriers respectively.}
            \label{fig:event-publishers}
\end{figure*}

The line graphs (see Figures \ref{fig:event-publishers}) compare the number of publishers, the average number of articles per publisher, and the average number of events per publisher for all the ten categories and the five barriers. Overall, it can be seen that the average articles and events per publisher are far higher in the political barrier for all ten categories, whereas the number of articles is far higher in the geographical barrier for all ten categories.

With regard to the number of publishers for all the barriers in all ten categories, there is a huge difference in the business category such as the number of publishers are almost double for linguistic barrier than the cultural, economic, and political barrier, and similarly, double than the linguistic barrier for the geographic barrier. Then there is fluctuation for all the categories after a straight decline at the category games. Overall, the noticeable fact from this diagram is that the linguistic barrier includes the highest number of news publishers. The other three barriers have small variations for all the categories.

The average number of news articles per publisher is almost equal for the economic and cultural barriers and the geographic and linguistic barriers. Whereas for the political barrier, it is always high for all the ten categories. We can see that the science category includes almost 280 news articles per publisher whereas in the health, home, and recreation categories, the count is almost 60 news articles per publisher and in business, shopping, and sports, the count is almost equal to 40 news articles per publisher. 

With regard to the number of events per publisher, the pattern is the same as the average number of news articles per publisher for the political, linguistic, economic, and cultural barriers. However, for the geographic barriers, this count reduces to almost half for the seven categories (business, computers, health, home, recreation, science, and society).

The popularity of events can be shown by the number of news articles published by different news publishers and the scope of a category can be depicted with coverage \cite{ref:sen2015clicks}. We can see that ten different categories have different scopes across different barriers. However, we notice that the science and society categories have the highest number of news publishers and the highest average number of news articles and news events per publisher for all the barriers whereas the games category appears with a scarcity of popularity. 

\textbf{Q2: What prominent relations appear between meta-data such as political alignment, geographical place, economic conditions, cultural values, and publishing language?}
Since the purpose of using semantic knowledge was to improve text classification, we analyzed the associated Wikipedia concepts to all the barriers. Also, we compared the occurrence of the list of Wikipedia concepts between the categories. 

We present an example to illustrate the comparison. To perform a comparison between all the barriers, we select the society category whereas to perform a comparison between the categories, we select the computers and society categories. The results of the intersection between the categories have been shown in Figure \ref{fig:venndiagrams}.

\section{Experimental Results}\label{sec:ExpResults}
\begin{figure}
\centering
\includegraphics[keepaspectratio=true,scale=0.5]{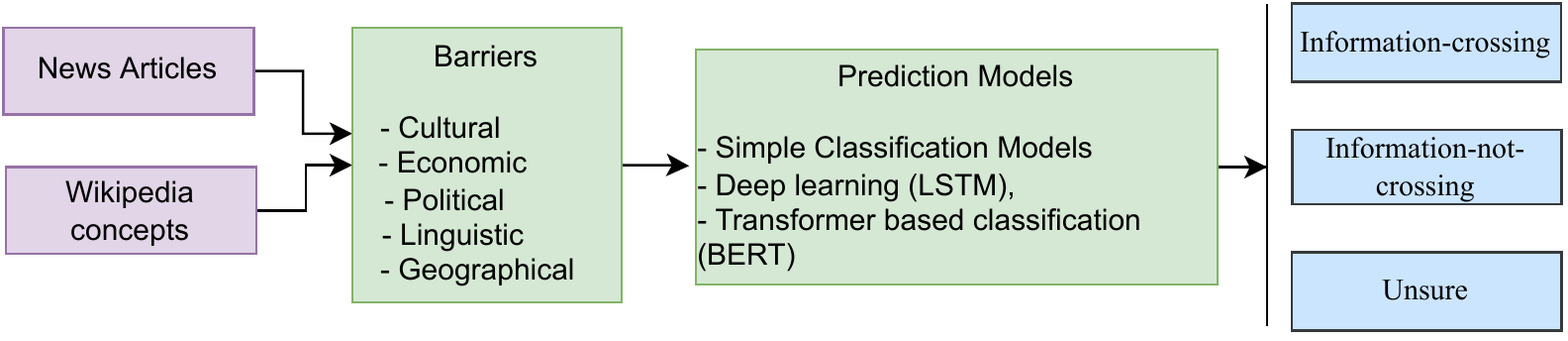}
\caption{Overview of the task of barrier classification using the Wikipedia concepts}
\label{fig:overview}
\end{figure}

In this section, we present an analysis of information spreading and Wikipedia concepts, classification baselines, evaluation metric, and experimental results comparing simple (LR, SVM, DT, RF, kNN), deep learning (LSTM), and transformers (BERT) for the barrier classification task (see Figure \ref{fig:overview}).

\begin{figure*}
            \centering
            \includegraphics[keepaspectratio=true,scale=0.36]{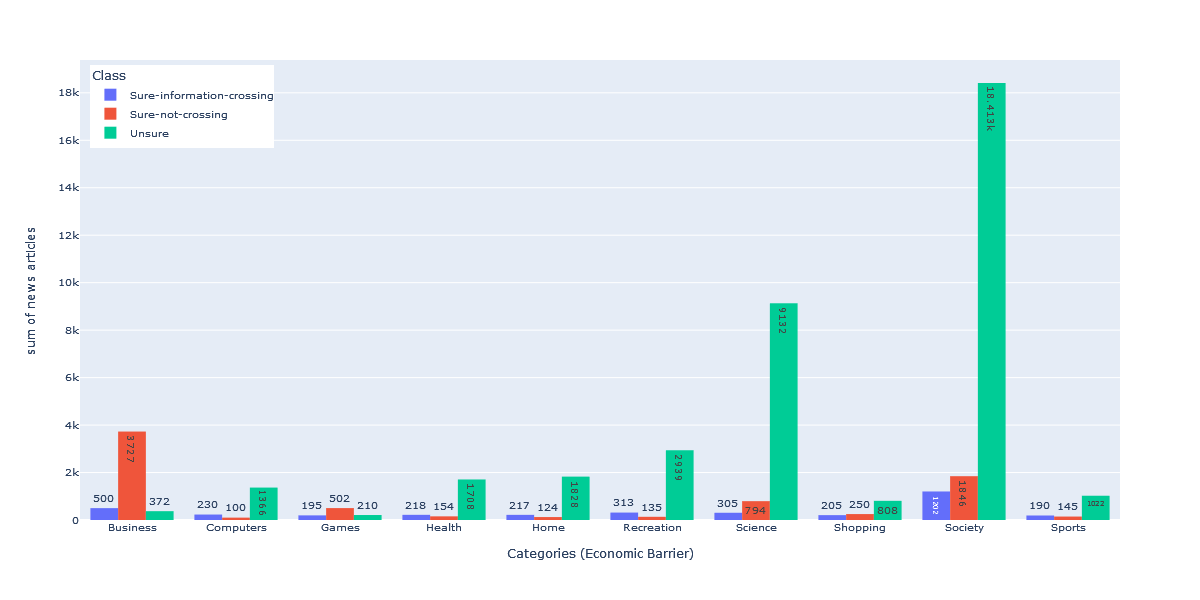}
            \includegraphics[keepaspectratio=true,scale=0.36]{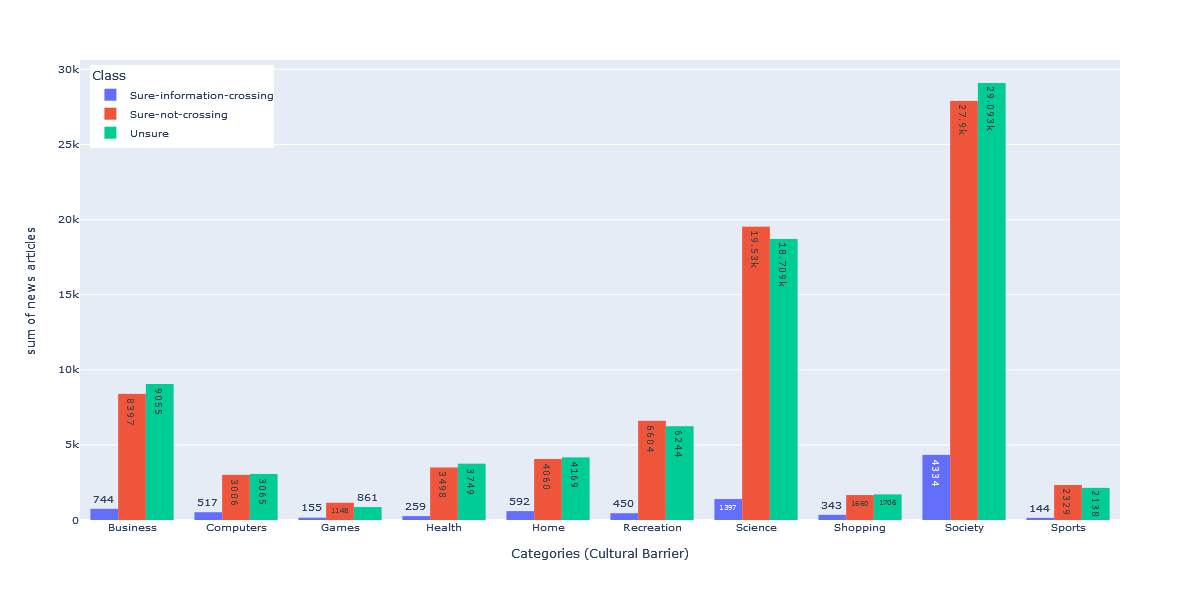}
            \caption{\bf This bar chart shows the class distribution for the economic, and cultural barriers (from left to right). The bar with red color shows the distribution for the class "Information-not-crossing" whereas the bar with green color shows the distribution for the class "Unsure" a barrier. The bar with blue color shows the distribution for the class "Information-crossing". Each of the two bar charts presents the class distribution for all ten categories.}
            \label{fig:ecoculDist}
\end{figure*}

\subsection{Evaluation Methodology}
We used Scikit-learn implementation of classical and deep learning models considering the following parameters, which are usually the default: hidden layers = 3, hidden units = 64, no. of epochs = 10, batch size = 64, and dropout = 0.001. For the training process of political, geographical, and linguistic barriers, we used Adam as the optimizer, categorical cross-entropy as the loss function, and sigmoid as the activation function. For economic and cultural barriers, we used Adam as the optimizer, binary cross-entropy as the loss function, and SoftMax as the activation function. 

\subsection{Baselines}
For the comparison with the proposed Wikipedia concepts based semantic knowledge, we evaluated the barrier classification task using the body text of the news articles only. We adopted the term frequency (TF) and inverted document frequency (IDF) methods to represent the bag of words of each news article. For the barrier classification task, the experiments were conducted by utilizing three different types of machine learning algorithms: 1) traditional machine learning algorithms including Logistic Regression (LR), Naive Bayes (NB), Support Vector Classifier (SVC), k-nearest Neighbor (kNN), and Decision Tree (DT): The performance of LR  for the text classification problems is same as of the SVM algorithm \cite{ref:shah2020comparative, ref:shah2020comparative}. SVMs use kernel functions to find separating hyper-planes in high-dimensional spaces \cite{ref:colas2006comparison}. SVM is difficult to interpret and there have to be many parameters that need to be set for performing the classification and one parameter that performs well in one task might perform poorly in other\cite{ref:shah2020comparative, ref:shah2020comparative}. Therefore many information retrieval systems use decision trees and naive bayes. However, these models lack accuracy \cite{ref:kowsari2017hdltex, ref:kamath2018comparative}. 2) LSTM (Long-Sort-term Memory): With the emergence of deep learning algorithms, the accuracy of text categorization has been greatly improved. Convolutional neural networks (CNN) and long short-term memory networks (LSTM) are widely used \cite{ref:luan2019research, ref:yu2020attention, ref:luan2019research, ref:kamath2018comparative, ref:wang2017comparisons}. 3) State-of-the-art pre-training language model BERT (Bidirectional Encoder Representations from Transformers): It is trained on a large network with a large amount of unlabeled data and adopts a fine-tuning approach that requires almost no specific architecture for each end task and has achieved great success in a couple of NLP tasks, such as natural language inference, and text classification \cite{ref:yu2019improving, ref:jin2020bert, ref:gonzalez2020comparing}.

\subsection{Evaluation metric}\label{subsec:evalutionmetric}
To evaluate the performance of binary and multi-class barrier classification models, Accuracy and F1-score is used as evaluation measure.
\begin{itemize}
  \item \textbf{F1-Score:} It combines the precision and recall of a classifier into a single metric by taking their harmonic mean.
  It is defined as:
                     \[ F_{1} =  \frac {2(Precision * Recall)}{Precision + Recall} \] 
\end{itemize}
\begin{itemize}
  \item \textbf{Accuracy:} Accuracy is a metric used in classification problems and it is used to tell the percentage of accurate predictions (TP and TN). We calculate it by dividing the number of correct predictions (TP and TN) by the total number of predictions (TP+FP+TN+FN).
  It is defined as:
                     \[ Accuracy{} =  \frac {TP+TN}{TP+FP+TN+FN} \] 
\end{itemize}

\subsection{Comparative analysis of the ten categories}\label{subsec:tencats}
We compare the results of all ten news categories based on evaluation metrics, i.e. accuracy, and F1-score. The both matrices are compared on the bar chart in order to display a concise and perfect comparison. Since the results of LR among the five (LR, SVC, NB, DT, and kNN) traditional machine learning algorithms were higher in all the categories, we exclude the others. The words PM-LSTM (proposed model LSTM) and PM-BERT (proposed model BERT) in the figure \ref{fig:wiki-cate-f1} mean the usage of LSTM and BERT utilizing our approach with Wikipedia concepts based semantic knowledge.

\subsubsection{F1-Score}
\begin{figure}
\centering
\includegraphics[keepaspectratio=true,scale=0.28]{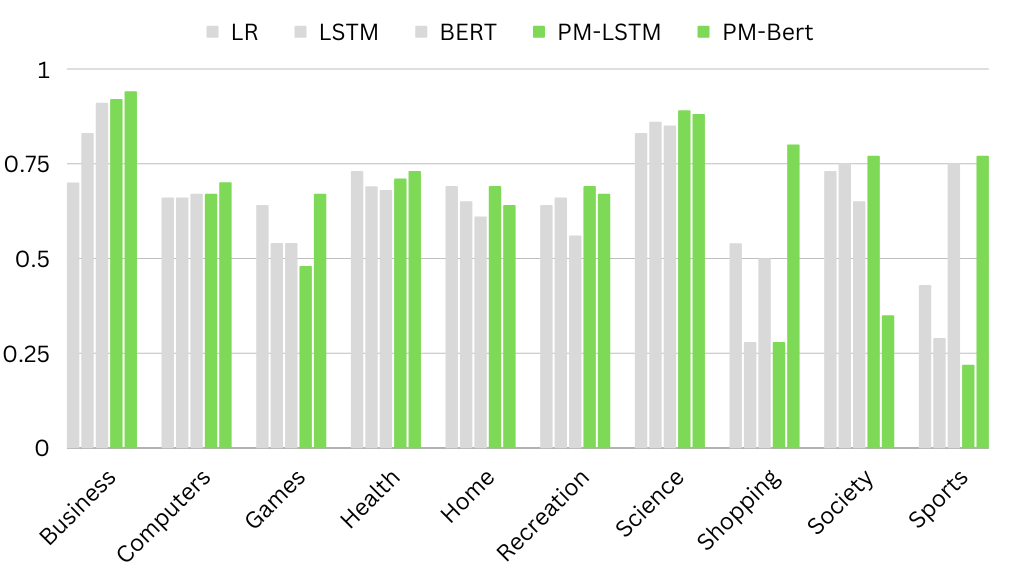}
\caption{It presents the F1 score of the five different machine learning algorithms (LR, LSTM, BERT, PM-LSTM, and PM-BERT) for the ten different categories (business, computers, games, health, home, recreation, science, shopping, society, and sports).}
\label{fig:wiki-cate-f1}
\end{figure}

The obtained bar chart is shown in Figure \ref{fig:wiki-cate-f1}.
It compares the results of LR, LSTM, and BERT with our proposed approach that is based on Wikipedia concepts based semantic knowledge. The F1 scores using BERT with the Wikipedia concepts based semantic knowledge are higher than LR, LSTM, and BERT for the business, computers, games, shopping, and sports (with the improvement of 0.03, 0.03, 0.03, 0.36, and 0.02 F1 score respectively); In case of recreation, science, and society, LSTM with our approach achieves higher F1 score (with the improvement of 0.03, 0.03, and 0.02 F1 score respectively); In case of health and home categories, we did not see any improvements of our approach in the results. 

\subsubsection{Accuracy}
\begin{figure}
\centering
\includegraphics[keepaspectratio=true,scale=0.28]{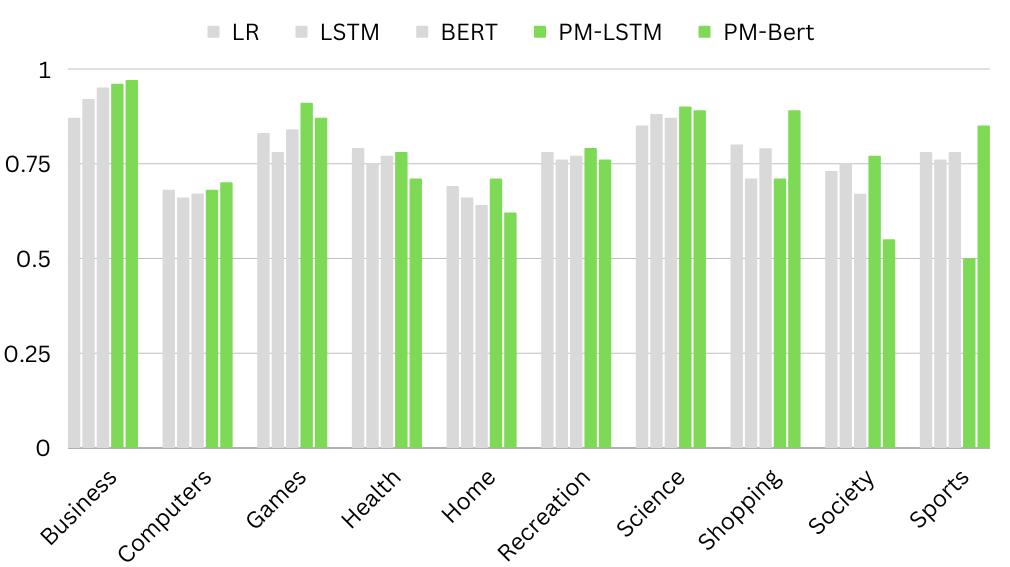}
\caption{It presents the accuracy of five different machine learning algorithms (LR, LSTM, BERT, PM-LSTM, and PM-BERT) for the ten different categories (business, computers, games, health, home, recreation, science, shopping, society, and sports).}
\label{fig:wiki-cate-acc}
\end{figure}

The obtained bar chart is shown in Figure \ref{fig:wiki-cate-acc}. The accuracy using LSTM with Wikipedia concepts based semantic knowledge is higher than LR, LSTM, and BERT for games, home, recreation, science, and society (with the improvement of 0.07, 0.02, 0.01, 0.02, and 0.02 accuracy score respectively); In case of business, computers, shopping, and sports categories, BERT model with our approach achieves higher accuracy (with the improvement of 0.02, 0.02, 0.09, and 0.07 accuracy score respectively);
By comparing and analyzing the results of different classification methods on ten different kinds of news categories, we can say that Wikipedia concepts based semantic knowledge helps in achieving a higher F1 score and accuracy.

\subsection{Comparative analysis of the three types of algorithms}\label{subsec:algos}
\begin{figure}
\centering
\includegraphics[keepaspectratio=true,scale=0.28]{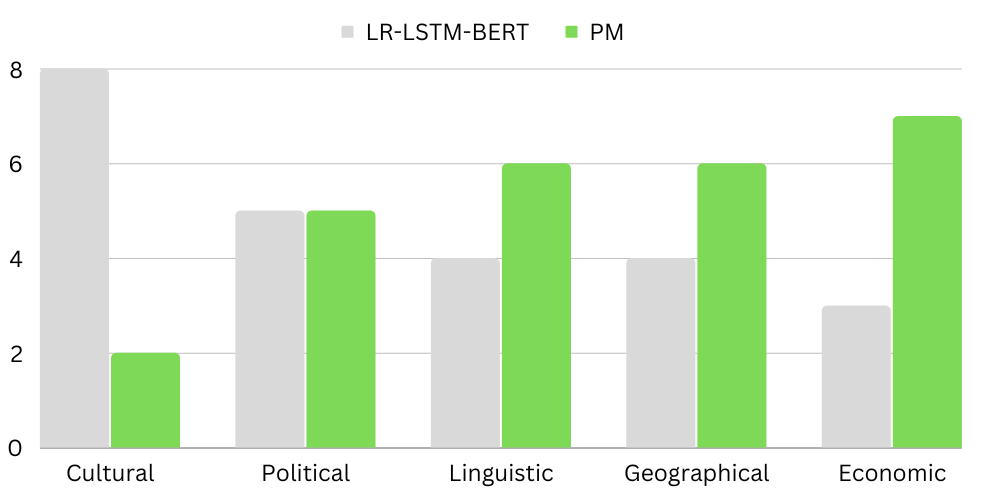}
\caption{It presents two bars for each barrier. The green bar means the number of categories for whom the classification methods show improved F1 and accuracy scores using our proposed approach (using Wikipedia concepts based semantic knowledge). The gray bar means the number of categories for whom the classification methods do not improve the F1 and accuracy score.}
\label{fig:barrier-results}
\end{figure}

After discussing the results of all the ten news categories, we compare all the five different types of barriers based on improvements in classification results. Figure \ref{fig:barrier-results} presents the statistics about each barrier. 
\newline
\textbf{Q3:} Which classification methods (classical or deep learning methods) yield the best performance to barrier classification task?
For the linguistic and geographic barrier, we see that our proposed methods (LSTM and BERT with semantic knowledge) outperform for six categories whereas for the five categories of political barrier, a slight improvement in classification results have been seen. It is also noticeable that the there are seven categories in economic barrier where proposed methods yields the best score. However, there are slight improvement for cultural barrier.

\subsection{Analysis and discussion}\label{subsec:AnalysisAndDiscussion}
Experiments of the novel approach on the ten different kinds of news and for the five different barriers have brought some insights regarding information spreading. In order to support the hypothesis, we have set three research questions \ref{subsec:hypothesis}. To answer the first research question (Does the information spreading in news varies across different topics and different barriers?), we compare the number of news publishers, the average number of articles per publisher, and the average number of events per publisher for all the categories and barriers (see Figure \ref{fig:event-publishers}). The comparative analysis indicates that the ten different categories have different scopes across the different barriers. However, the society and science categories appeared to have the highest number of news publishers, the highest average number of news articles, and the news events per publisher for all the barriers whereas the games category appeared to have a minimum number of articles and publishers. To answer the second research question (What prominent relations appear between Wikipedia concepts, and different barriers and categories?), we find the intersection between the Wikipedia concepts belonging to different barriers and categories (see Figure \ref{fig:venndiagrams}). The results suggest that although Wikipedia concepts are shared among the barriers, a category in each barrier has some unique Wikipedia concepts. Similarly, the same fact exists between the different categories. Therefore it might be possible that it will help in improving the classification results. The results of the annotation show that the data does not have higher imbalanced data for both binary and ternary class classification (see Figures \ref{fig:barriDistclass}, \ref{fig:ecoculDist}). Therefore we consider using it for classification without using any technique to make it balanced. To answer our third research question (Which classification methods (classical or deep learning methods) yield the best performance to barrier classification task?), We perform classification with traditional machine learning methods including Logistic Regression (LR), Naive Bayes (NB), Support Vector Classifier (SVC), k-nearest Neighbor (kNN), and Decision Tree (DT). Afterward, we perform classification with and without Wikipedia concepts using LSTM and BERT. We evaluate the models using accuracy and F1 score (see Subsection \ref{subsec:evalutionmetric}). We analyze the classification results by comparing the ten categories \ref{subsec:tencats} and three types of classification methods \ref{subsec:algos}. The results suggest that for the linguistic and geographic barrier, our proposed approach yields the best scores for the six categories, whereas for the political barrier, we see a slight improvement in the classification of the five categories. On the other hand, LSTM and BERT with Wikipedia concepts yield the best score for the seven categories of the economic barrier. Overall, we can say that Wikipedia concepts-based semantic knowledge help in achieving a higher F1 score and accuracy.

\section{Conclusions}\label{sec:conclusions}
In this paper, we focused on the classification of news-spreading barriers by utilizing semantic knowledge in form of Wikipedia concepts. We consider news related to ten different categories (business, computers, games, health, home, recreation, science, shopping, society, and sports). After completing the automatic annotation of news data for the five barriers including cultural, economic, political, linguistic, and geographical (binary class classification of the linguistic, political, and geographical barrier and ternary class classification of the cultural and political barrier), we perform classification with traditional machine learning methods (LR, NB, SVC, kNN, and DT), deep learning (LSTM) and transformer-based method (BERT). Our findings suggest that Wikipedia concepts-based semantic knowledge help in achieving a higher F1 score and accuracy.
\newline

\section{Acknowledgments}\label{sec:ack}
The research described in this paper was supported by the Slovenian research agency under the project J2-1736 Causalify and by the European Union’s Horizon 2020 research and innovation program under the Marie Skłodowska-Curie grant agreement No 812997.
\FloatBarrier

\bibliographystyle{plain}
\bibliography{Main.bib}   

\end{document}